\documentclass[10pt,journal,compsoc]{IEEEtran}

\ifCLASSOPTIONcompsoc
  \usepackage[nocompress]{cite}
\else
  \usepackage{cite}
\fi

\ifCLASSINFOpdf
\else
\fi

\usepackage{hyperref}
\usepackage{amsmath}
\usepackage{xcolor} 
\usepackage{mathtools}
\usepackage{algorithm}
\usepackage{algpseudocode}
\usepackage{wrapfig,lipsum,booktabs}
\usepackage{xcolor}
\usepackage{amssymb}
\usepackage{pgf-pie}
\usepackage{capt-of}
\usepackage{float}
\usepackage{svg}
\colorlet{shadecolor20}{gray!20}
\colorlet{shadecolor10}{gray!10}

\usepackage[dvipsnames]{xcolor}

\usepackage[capitalize]{cleveref}
\crefname{figure}{Fig.}{Figs.}
\Crefname{figure}{Figure}{Figures}
\crefname{section}{Sec.}{Secs.}
\Crefname{section}{Section}{Sections}
\Crefname{table}{Table}{Tables}
\crefname{table}{Tab.}{Tabs.}
\Crefname{paragraph}{Paragraph}{Paragraphs}
\crefname{paragraph}{Para.}{Paras.}

\definecolor{softblue}{RGB}{173, 216, 230}   
\definecolor{softgreen}{RGB}{144, 238, 144}  
\definecolor{softpink}{RGB}{255, 182, 193}   
\definecolor{softgray}{RGB}{211, 211, 211}   

\definecolor{forward}{RGB}{165,93,115}

\def\renderer{\textcolor{forward}{\texttt{renderer}}}

\algrenewcommand\algorithmicrequire{\textbf{Input:}} 
\algrenewcommand\algorithmicensure{\textbf{Output:}}

\newcommand{\methodName}{Gen-3Diffusion}

\colorlet{shadecolor20}{gray!20}
\colorlet{shadecolor10}{gray!10}

\begin{document}
\title{Gen-3Diffusion: Realistic Image-to-3D Generation via 2D \& 3D Diffusion Synergy}

\author{\fontsize{10}{13.6}\selectfont{Yuxuan~Xue\textsuperscript{{1,2}} 
        Xianghui~Xie\textsuperscript{{1,2,3}}
        Riccardo~Marin\textsuperscript{{1,2}}
        Gerard Pons-Moll\textsuperscript{{1,2,3}}
        \thanks{E-mail:yuxuan.xue@uni-tuebingen.de} \\
        \vspace{0.2cm}\textsuperscript{{1}}University of T{\"u}bingen \quad \textsuperscript{{2}}T{\"u}bingen AI Center \quad \textsuperscript{{3}}Max Planck Institute for Informatics, Saarland Informatics Campus
        }
        }%

\markboth{IEEE TRANSACTIONS ON PATTERN ANALYSIS AND MACHINE INTELLIGENCE, VOL. 0, NO. 0, 2025}%
{Shell \MakeLowercase{\textit{et al.}}: Bare Demo of IEEEtran.cls for Computer Society Journals}

\IEEEtitleabstractindextext{%
\begin{abstract}
Creating realistic 3D objects and clothed avatars from a single RGB image is an attractive yet challenging problem. 
Due to its ill-posed nature, recent works leverage powerful prior from 2D diffusion models pretrained on large datasets.
Although 2D diffusion models demonstrate strong generalization capability, they cannot guarantee the generated multi-view images are 3D consistent.
In this paper, we propose \textbf{Gen-3Diffusion}: Realistic Image-to-3D Generation via 2D \& 3D Diffusion Synergy.
We leverage a pre-trained 2D diffusion model and a 3D diffusion model via our elegantly designed process that synchronizes two diffusion models at both training and sampling time.
The synergy between the 2D and 3D diffusion models brings two major advantages: 
1) \textbf{2D helps 3D in generalization}: the pretrained 2D model has strong generalization ability to unseen images, providing strong shape priors for the 3D diffusion model; 
2) \textbf{3D helps 2D in multi-view consistency}: the 3D diffusion model enhances the 3D consistency of 2D multi-view sampling process, resulting in more accurate multi-view generation.
We validate our idea through extensive experiments in image-based objects and clothed avatar generation tasks. Results show that our method generates realistic 3D avatars and objects with high-fidelity geometry and texture. Extensive ablations also validate our design choices and demonstrate the strong generalization ability to diverse clothing and compositional shapes. 
Our code and pretrained models will be publicly released on our \href{https://yuxuan-xue.com/gen-3diffusion}{project page}.

\end{abstract}

}

\maketitle

\IEEEdisplaynontitleabstractindextext

\IEEEpeerreviewmaketitle

\section{Introduction}
\label{sec:introduction}

\begin{figure*}
\includegraphics[width=1.0\textwidth]{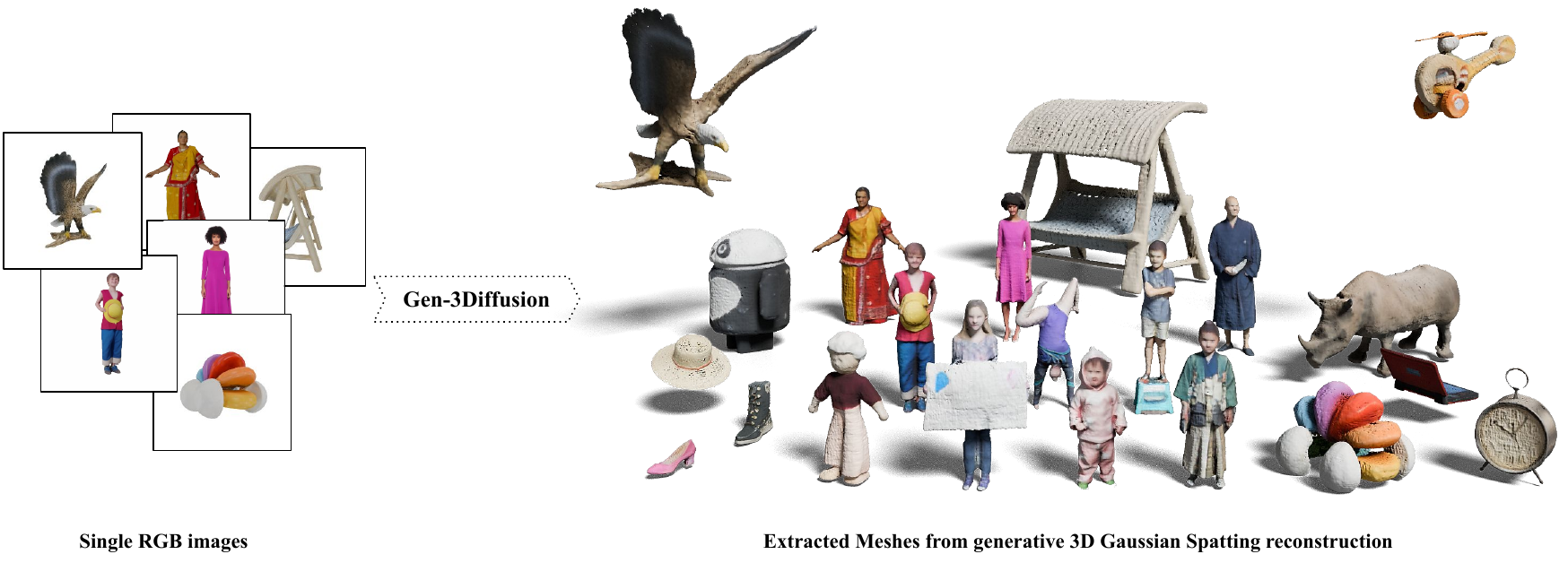}
\vspace{-0.5cm}
\caption{Given a single image of a person or an object, our method \textbf{Gen-3Diffusion} creates realistic 3D objects or clothed avatars with high-fidelity geometry and texture. We use Gaussian Splatting to flexibly represent various shapes which can be extracted to high-quality textured meshes.}
\label{fig:teaser}
\end{figure*}
\IEEEPARstart{C}{reating} realistic 3D content is crucial for numerous applications, including AR/VR, as well as in the movie and gaming industries. Methods for creating a 3D model from a single RGB image are especially important to scale up 3D modelling and make it more consumer-friendly compared to traditional studio-based capture methods. However, this task presents substantial challenges due to the extensive variability in object shapes and appearances. These challenges are further intensified by the inherent ambiguities associated with monocular 2D views.

Moreover, beyond general object modeling, the generation of realistic clothed avatars presents a particularly demanding set of challenges. This complexity arises from the diversity of human body shapes and poses, which is compounded by a wide array of clothing, accessories, and occlusion by interacting objects. Such challenges are accentuated by the relative scarcity of large-scale 3D human datasets as compared to those available for objects, highlighting the critical need for advanced modeling techniques that can effectively navigate these complexities.

\begin{figure*}[htbp]
    \centering
    \begin{minipage}[b]{0.49\textwidth}
        \centering
        \includegraphics[width=\textwidth]{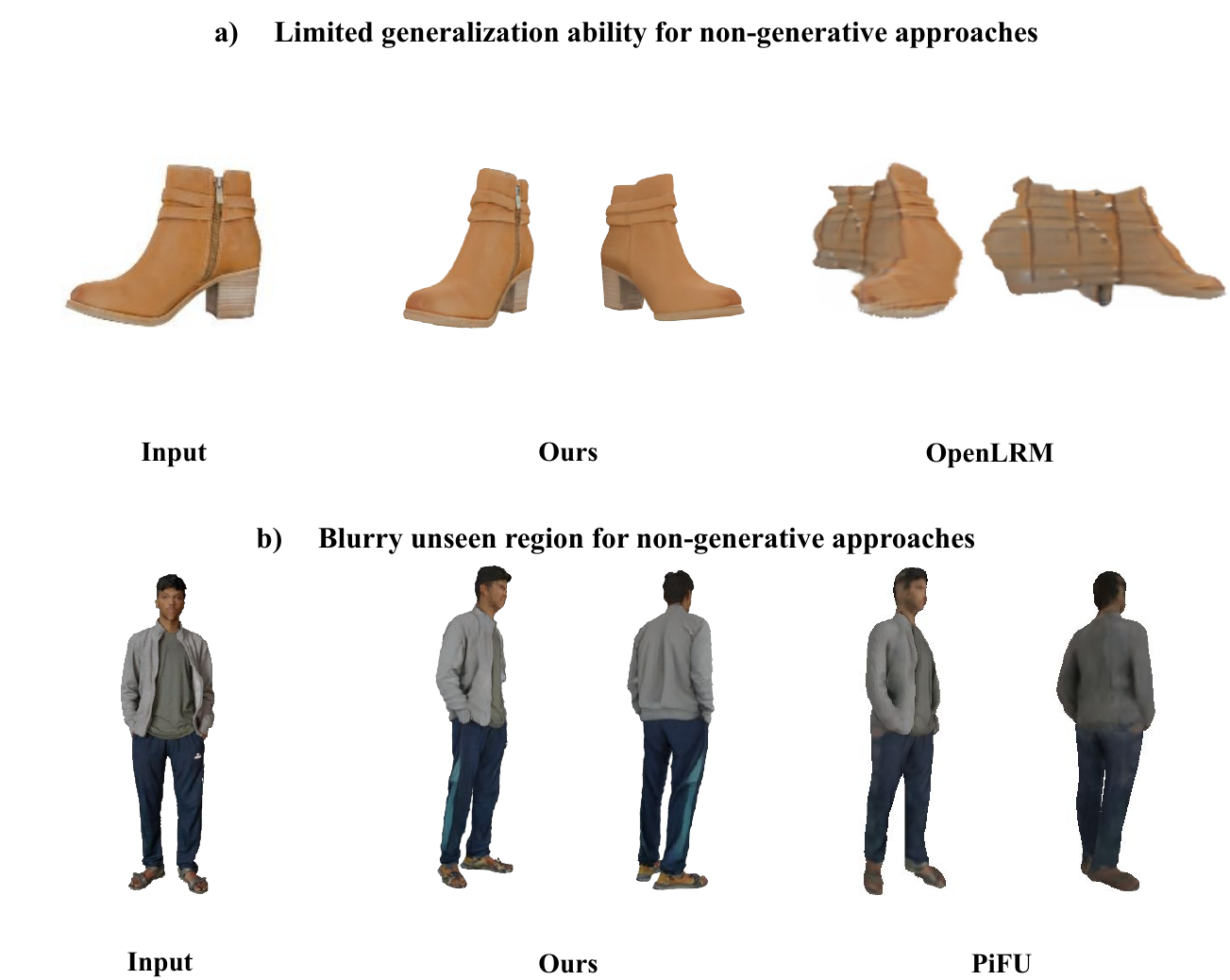} 
        \vspace{-0.5cm}
        \caption{\textbf{Motivation for generative 3D reconstruction design}. Unlike methods~\cite{hong2023lrm, saito2019pifu} that deterministically regress 3D from single images, our \methodName{} learns conditional distribution and samples a plausible 3D-GS, resulting in high-fidelity and realistic unseen regions.
        }
        \label{fig:object_motivation}
    \end{minipage}
    \hfill
    \begin{minipage}[b]{0.49\textwidth}
        \centering
        \includegraphics[width=\textwidth]{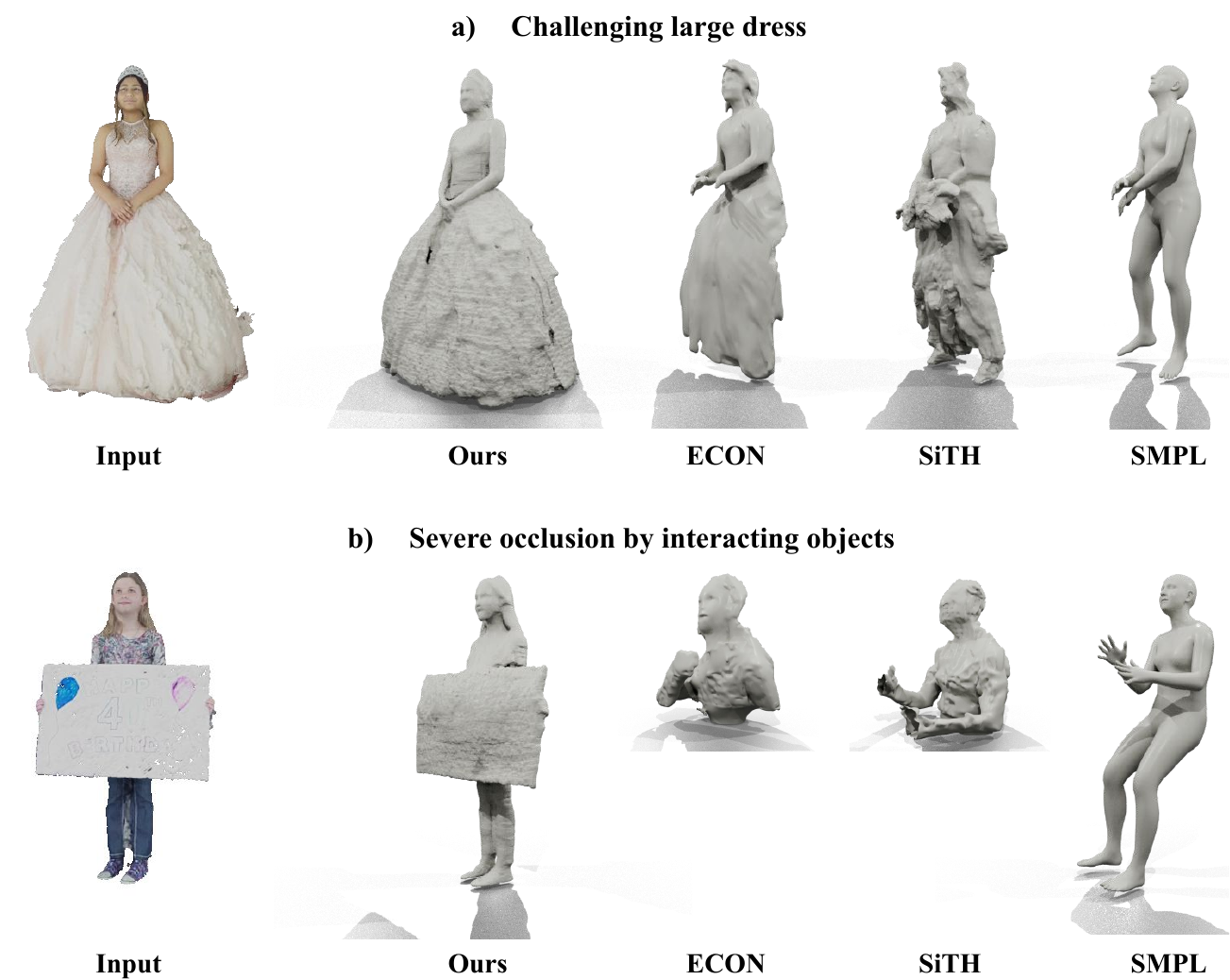} 
        \vspace{-0.5cm}
        \caption{\textbf{Motivation for template-free avatar reconstruction design}. Methods~\cite{xiu2023econ, ho2023sith} relying on SMPL~\cite{loper2015smpl} template suffer from inaccurate SMPL estimation and cannot represent challenging dresses or object interaction. Our \methodName{} is template-free and leverages shape prior from 2D diffusion models, can faithfully handle above challenges.
        }
        \label{fig:human_motivation}
    \end{minipage}
\end{figure*}

Recent image-to-3D approaches can be categorized into Direct-Reconstruction-based and Multi-View Diffusion-based methods. 
Direct-Reconstruction-based approaches directly predict a 3D representation that can be rendered from any viewpoint. Due to the explicit 3D representation, these methods produce an arbitrary number of consistent viewpoint renderings. 
For objects reconstruction, recent approaches such as LRM~\cite{hong2023lrm} and TriplaneGaussian~\cite{zou2023triplanegaussian} directly predict the NeRF or 3D Gaussian Splats from the input context view.
However, these non-generative models directly regress the 3D representation in a deterministic manner, which easily leads to blurry unseen regions in~\cref{fig:object_motivation}.
For clothed avatar reconstruction, recent popular approaches obtain the 3D model based on common template ~\cite{ho2023sith, xiu2023icon, xiu2023econ, zhang2023sifu} which utilize the SMPL~\cite{loper2015smpl} body model as the shape prior and perform the clothed avatar reconstruction.
However, underlying SMPL body template highly limits the 3D representation of challenging human appearance, such as large dress, occlusion by interacting objects, etc. Examples can be found in~\cref{fig:human_motivation}.
Furthermore, human reconstructors are trained on relative small-scale datasets due to the limited amount of high-quality 3D human data, which further restricts their ability to generalize to diverse shapes and textures.
Last but not least, all above mentioned image-to-3D reconstruction works, regardless object-oriented or human-oriented, are typically deterministic which produce blurry textures and geometry in the occluded regions. 

Multi-view diffusion-based methods~\cite{Liu2023zero123, shi2023zero123++, Wang2023ImageDream} are proposed to synthesize desired novel views from single RGB image. These methods distill the inherent 3D structure presented in pretrained 2D diffusion models~\cite{Rombach2022StableDiffusion}.
Typically, they fine-tune a large-scale 2D foundation model~\cite{Rombach2022StableDiffusion} on a large 3D dataset of objects~\cite{Deitke2023Objaverse, wu2023omniobject3d, yu2023mvimagenet}, to generate novel views at given camera poses. Thanks to the pertaining on large-scale image datasets, Multi-view diffusion methods show strong generalization capability to unseen objects. However, since these models diffuse images purely in 2D without explicit 3D constraints or representation, the resulting multi-views often lack 3D consistency~\cite{qian2023magic123, liu2023one2345}. The 3D inconsistent multi-view images further restrict downstream applications such as sparse-view 3D reconstruction~\cite{Tang2024LGM}.

To address these challenges, we propose \textbf{\methodName}: Realistic Image-to-3D Generation via 2D \& 3D Diffusion Synergy. 
We design our method based on two key insights: 
1). 2D multi-view diffusion models (MVDs) provide strong shape priors that help 3D reconstruction; 
2). Explicit 3D representation produces guaranteed 3D consistent multi-views that improve the accuracy of sampled 2D multi-view images. 
To leverage the benefits of both 2D MVD and explicit 3D representation, we propose a novel framework that synchronizes a 3D diffusion model with a 2D MVD model at each diffusion sampling step. 

Specifically, we first introduce a novel 3D diffusion model that directly regresses 3D Gaussian Splatting (3D-GS~\cite{Kerbl20233dgs}) from intermediately denoised multi-views images of 2D MVD. The predicted 3D-GS can be rendered into multi-view images with guaranteed 3D consistency. At every iteration, 2D MVD denoises multi-view images conditioned on input view, which are then reconstructed to 3D-GS by our 3D diffusion model. The 3D-GS are then re-rendered to multi-views to continue the diffusion sampling process.  
This 3D lifting during iterative sampling improves the 3D consistency of the generated 2D multi-view images while leveraging a large-scale foundation model trained on billions of images. 

In summary, our contributions are: 
\begin{itemize}
    \item We propose a novel 3D-GS diffusion model for 3D reconstruction, which bridges large-scale priors from 2D multi-view diffusion models and the efficient explicit 3D-GS representation.
    \item A sophisticated joint diffusion process that incorporates reconstructed 3D-GS to improve the 3D consistency of 2D diffusion models by refining the reverse sampling trajectory.   
    \item Our proposed formulation enables us to achieves superior performance and generalization capability than prior works, both in fields of objects reconstruction ($\textit{Gen3D}_\text{object}$) and clothed human reconstruction ($\textit{Gen3D}_\text{avatar}$). 
    Our code and pretrained models will be publicly released on our \href{https://yuxuan-xue.com/gen-3diffusion}{project page}.

\end{itemize}

\section{Related Works}
\label{sec:related_works}

\subsection{Novel View Synthesis}
Significant progress has been made in recent years in synthesizing images at target camera poses given multi-view observations. NeRF~\cite{Mildenhall2020NeRF} and 3D Gaussian splitting (3D-GS)~\cite{Kerbl20233dgs} are two popular representations for novel view synthesis. 
NeRF~\cite{Mildenhall2020NeRF} uses neural networks to represent the continuous radiance fields and obtains new images via volumetric rendering. Despite impressive results, the training and rendering speed is slow and lots of efforts~\cite{mueller2022instantngp, Chen2022ECCV_tensoRF} have been made to speed up NeRF. Alternatively, 3D-GS~\cite{Kerbl20233dgs} represents the radiance with a discrete set of 3D Gaussians and renders them with rasterization which is highly efficient. 

Optimizing NeRF or 3D-GS is time-consuming and requires dense multi-view images. Recently, Zero-1-to-3~\cite{Liu2023zero123} proposes a novel idea in fine-tuning pretrained image diffusion models~\cite{Rombach2022StableDiffusion} to generate the desired target views in a zero-shot manner. Considering the power of the pretrained StableDiffusion~\cite{Rombach2022StableDiffusion}, Zero-1-to-3 has seen 5B 2D images and 10M 3D objects, demonstrating superior generalization ability in real-world images. 

Despite the excellent generalization ability, zero-1-to-3 suffer from severe 3D inconsistency across different views while generating multiple different views. The reason is that the different views are sampled independently from each other. 
To address the multi-view inconsistency in diffusion sampling, multi-view diffusion models~\cite{long2023wonder3d, shi2023zero123++, Wang2023ImageDream, liu2023syncdreamer} are proposed to generate multiple views simultaneously with information exchange across all sampled views using dense pixel-level attention in the latent space. In our observation, the dense multi-view attention improves the multi-view consistency, but also limits the ability of free novel view synthesis in Zero-1-to-3. Moreover, the generated multi-view images still have no guarantee of 3D consistency due to the lack of a common 3D representation during the diffusion sampling process. 

\subsection{3D Objects Generation}
Obtaining high-quality 3D objects from a single Image is an attractive but challenging task.
Early object reconstruction works~\cite{Alwala_CVPR22_pretrain_ss, thai3dv2020SDFNet, shapehd, Xian2022gin, genre} focus mainly on geometry.
With the emergence of differentiable rendering technologies, many works try to directly regress a 3D representation such as NeRF~\cite{Yu2021pixelnerf, hong2023lrm} or 3D-GS~\cite{zou2023triplanegaussian, charatan2024pixelsplat} from single RGB images. However, these methods are deterministic and do not learn the distribution of the underlying 3D scene, which can result in blurry rendering results at the inference time as in~\cref{fig:object_motivation}. 
In parallel to direct regression of 3D from single images, a line of work explores learning unconditional or conditional 3D generative models from multi‐view data by integrating differentiable rendering into training~\cite{Tewari2023DiffusionWithForward, cai2024bakegs}. These approaches can synthesize both object‐ and scene‐level geometry and appearance. However they do not utilize the prior of large, pre‐trained 2D models, which can cause slow convergence and require extensive compute to achieve high fidelity.
More recently, several methods aim to sidestep these issues by first encoding 3D assets into a compact VAE latent space and then training diffusion or transformer‐based generators therein~\cite{xiang2024trellis, he2025triposf, zhang2024clay}. While this two‐stage strategy yields impressive generation quality, the need to pretrain and optimize high‐capacity 3D VAEs makes it prohibitively expensive for most academic labs.

With the advance of 2D diffusion models~\cite{Rombach2022StableDiffusion} and efficient 3D representation~\cite{chan2022eg3d}, recent works can reconstruct 3D objects with detailed textures~\cite{hong2023lrm,liu2023one2345, long2023wonder3d, shi2023zero123++,  Tang2024LGM, tochilkin2024triposr, xu2024instantmesh, Xu2023DMV3D, zou2023triplanegaussian}. 
One popular paradigm is first using strong 2D models~\cite{Liu2023zero123, shi2023mvdream, Wang2023ImageDream} to produce multi-view images and then train another model to reconstruct 3D from multi-view images~\cite{liu2023one2345, liu2023one2345++, long2023wonder3d, Tang2024LGM, xu2024grm}. In practice, their performance is limited by the accuracy of the multi-view images generated by 2D diffusion modes. Some works have tried to train another network that learns to correct the noisy multi-view images~\cite{liu2023one2345, liu2023one2345++, xu2024instantmesh}. However, the network can be overfitted to error patterns from specific models, leading to limited generalization ability. Instead of correcting the multi-views in the last step output which is too late, we inject 3D consistency information early in the sampling stage, resulting in more accurate multi-views and 3D reconstruction. 
BiDiff~\cite{ding2024bidiff} introduces a novel dual‐branch diffusion that learns an SDF via 3D diffusion to guide 2D multi‐view diffusion network. However, BiDiff resamples noise directly on that 2D network output, where noise sampling step itself remains unconstrained and can break view consistency. 
In contrast, our approach embeds a fast 3D‐GS renderer directly into every diffusion iteration, so that geometric consistency is enforced at the pixel‐level throughout the entire sampling process, generating more 3D consistent multi‐view images at every step.


\subsection{Clothed Avatar from Image}
Creating realistic human avatar from consumer grade sensors~\cite{kabadayi24ganavatar, kim2024paintit,tiwari2021neuralgif, xue2022e-nr, xue2023nsf, xue2024e-nr-ijcv, xue2025infinihuman, xue2025physic} is essential for downstream tasks such as human behaviour understanding~\cite{bhatnagar22behave, petriv2023objectpopup, xie2022chore, xie2023template_free, xie2023vistrack} and gaming application~\cite{li2023diffavatar, liu2023gshell,guzov2021hps, zhang2022couch, zhang2024force}. Researchers have explored avatar creation from monocular RGB~\cite{jiang2022instantavatar, weng_humannerf_2022_cvpr}, Depth~\cite{dong2022pina, xue2023nsf} video or single image~\cite{saito2019pifu, saito2020pifuhd, sengupta_diffhuman_2024, xiu2023icon, xiu2023econ}.

Avatar from single image is particularly interesting and existing methods can be roughly categorized as template-based~\cite{ho2023sith, xiu2023icon, xiu2023econ, zhang2023sifu} and template-free~\cite{saito2019pifu, saito2020pifuhd, sengupta_diffhuman_2024, yang_d-if_2023}. 
Template-free approaches~\cite{saito2019pifu, saito2020pifuhd, sengupta_diffhuman_2024, yang_d-if_2023} directly predict a human occupancy field conditioned on a single image. This is flexible to represent diverse human clothing yet not robust to challenging poses due to the lack of shape prior. 
To leverage the shape prior information from human body models~\cite{loper2015smpl, pavlakos2019smplx}, template-based approaches first estimate parametric body mesh from the image and then reconstruct the clothed avatar.
Despite the impressive performance, these methods rely on the naked body model~\cite{loper2015smpl, pavlakos2019smplx} and they are affected by the inaccurate body mesh estimation which is common in extremely loose clothing or occlusion introduced by interacting objects as shown in~\cref{fig:human_motivation}. 
In this work, instead of using naked body models, we leverage the shape prior from pre-trained image diffusion models. This allows us to represent diverse clothed avatar shapes, including large dress and interacting objects. Furthermore, our method is not limited by the errors from monocular body mesh estimation methods. 

\section{Background}
\label{sec:background}

\subsection{Denoising Diffusion Probabilistic Models}
\label{sec:ddpm}
DDPM~\cite{Ho2020DDPM} is a generative model which learns a data distribution by iteratively adding (forward process) and removing (reverse process) the noise.
Formally, the forward process iteratively adds noise to a sample $\mathbf{x}_0$ drawn from a distribution $p_\text{data}(\mathbf{x})$:
\begin{equation}
\begin{gathered}
    \mathbf{x}_t \sim \mathcal{N}(\mathbf{x}_t; \sqrt{\alpha_t} \mathbf{x}_{t-1}, (1-\alpha_t)\mathbf{I}) = \sqrt{\bar{\alpha}_t}\mathbf{x}_0 + \sqrt{1-\bar{\alpha}_t}\mathbf{\epsilon}, \\ \text{ where } \mathbf{\epsilon}\sim\mathcal{N}(0, \mathbf{I}),
\end{gathered}
\label{eq:ddpm_forward}
\end{equation}
where $\alpha_t, \bar{\alpha}_t$ schedules the amount of noise added at each step $t$~\cite{Ho2020DDPM}. To 
sample data from the learned distribution,
the reverse process starts from $\mathbf{x}_T\sim\mathcal{N}(0, \mathbf{I})$ and iteratively denoises it until $t=0$: 
\begin{equation}
\begin{gathered}
    \mathbf{x}_{t-1}\sim\mathcal{N}(\mathbf{x}_{t-1};\mathbf{\mu}_\theta(\mathbf{x}_t, t), \tilde{\beta}_{t-1}\mathbf{I}), \\ \text{ where }\tilde{\beta}_{t-1}=\frac{1-\bar{\alpha}_{t-1}}{1-\bar{\alpha}_{t}} (1-\alpha_t)
    \label{eq:ddpm_reverse}
\end{gathered}
\end{equation}
A network parametrized by $\theta$ is trained to estimate the posterior mean $\mathbf{\mu}_\theta$ at each step $t$. One can also model conditional distribution with DDPM by adding the condition to the network input~\cite{dhariwal2021diffusionClassGuidance, ho2021classifierfreeCFG}. 

\subsection{2D Multi-View Diffusion Models}
\label{sec:mvd}
Many recent works~\cite{Liu2023zero123, liu2023syncdreamer, long2023wonder3d, shi2023zero123++, tang2023mvdiffusion++, Wang2023ImageDream} propose to leverage strong 2D image diffusion prior~\cite{Rombach2022StableDiffusion} pretrained on billions images~\cite{Schumann2022Laion5B} to generate multi-view images from a single image. 
Among them, ImageDream~\cite{Wang2023ImageDream} demonstrated a superior generalization capability to unseen objects~\cite{Tang2024LGM}.
Given a single context view image $\mathbf{x}^\text{c}$, ImageDream generate 4 orthogonal target views $\mathbf{x}^\text{tgt}$ with a model $\mathbf{\epsilon}_\theta$, which is trained to estimate the noise added at each step $t$. 
With the estimated noise $\mathbf{\epsilon}_\theta$, one can compute the "clear" target views $\Tilde{\mathbf{x}}^\text{tgt}_0$ with close-form solution in~\cref{eq:ddpm_forward}:  
\begin{align}
    \tilde{\mathbf{x}}_{0}^{\text{tgt}} &= \frac{1}{\sqrt{\bar{\alpha}_t}}( \mathbf{x}^{\text{tgt}}_t-\sqrt{1-\bar{\alpha}_t} \boldsymbol{\epsilon}_{\theta}(\mathbf{x}^{\text{tgt}}_t, \mathbf{x}^{\text{c}}, y, t)).
    \label{eq:one-step-mvd} 
\end{align}
This \emph{one-step} estimation of $\tilde{\mathbf{x}}_{0}^{\text{tgt}}$ can be noisy and inaccurate, especially when $t$ is large and $\mathbf{x}_t^{\text{tgt}}$ is extremely noisy and does not contain much information. 
Thus, the iterative sampling of $\mathbf{x}_{t}^{\text{tgt}}$ is required until $t=0$. To sample next step $\mathbf{x}_{t-1}^{\text{tgt}}$, standard DDPM~\cite{Ho2020DDPM} computes the posterior mean $\mu_\theta$ from current $\mathbf{x}_{t}^{\text{tgt}}$ and estimated $\tilde{\mathbf{x}}_{0}^{\text{tgt}}$ at step $t$ with:
\begin{equation}
\begin{gathered}
\begin{split}
    \mu_\theta(\mathbf{x}_t^\text{tgt}, t) &:=\mu_{t-1}(\mathbf{x}_t^\text{tgt}, \tilde{\mathbf{x}}_0^\text{tgt}) \\
    &= \frac{\sqrt{\alpha_{t}}\left(1-\bar{\alpha}_{t-1}\right)}{1-\bar{\alpha}_{t}} \mathbf{x}^{\text{tgt}}_{t} + \frac{\sqrt{\bar{\alpha}_{t-1}} \beta_{t}}{1-\bar{\alpha}_{t}}  \tilde{\mathbf{x}}_{0}^{\text{tgt}},
    \label{eq:xt_minus_one_from_xt_x0_old}
\end{split}
\\ \text{ where } \beta_t=1-\alpha_t.
\end{gathered}
\end{equation}
Afterwards, $\mathbf{x}_{t-1}^{\text{tgt}}$ can be sampled from Gaussian distribution with mean $\mu_{t-1}$ and variance $\tilde{\beta}_{t-1}\mathbf{I}$ (\cref{eq:ddpm_reverse}) and used as the input for the next iteration. The reverse sampling is repeated until $t=0$ where 4 clear target views are generated.

Although multi-view diffusion models~\cite{liu2023syncdreamer, shi2023zero123++, Wang2023ImageDream} generate multiple views together, the 3D consistency across these views is not guaranteed due to the lack of an explicit 3D representation. 
Thus, we propose a novel 3D consistent diffusion model, which ensures the multi-view consistency at each step of the reverse process by diffusing 2D images using reconstructed 3D Gaussian Splats~\cite{Kerbl20233dgs}.

\subsection{3D Diffusion with Differentiable Rendering}
Although DDPM~\cite{Ho2020DDPM} has emerged as a powerful class of generative models capable of capturing the distributions of complex signals, it can only model distributions for which training samples are
directly accessible.
Thus, directly training DDPM to learn the distribution of NeRF or 3D-GS requires pre-computing feature planes or Gaussians from 3D object scans, which is exorbitant. 
Recent works ~\cite{anciukevicius2023renderdiffusion, Tewari2023DiffusionWithForward, karnewar2023holodiffusion} propose to learn the distribution of 3D representation by diffusing the rendered images through differentiable rendering. 
In contrast to novel view diffusion models in~\cref{sec:mvd}, these works directly learn image-conditional 3D radiance field generation, instead of sampling from the distribution of novel views conditioned on a context view. 

Given a single context view image $\mathbf{x}^\text{c}$, Diffusion-with-Forward (DwF)~\cite{Tewari2023DiffusionWithForward} generates Pixel-NeRF~\cite{Yu2021pixelnerf} from the noisy view $\hat{\mathbf{x}}_{t}^\text{tgt}$ and render to clear target view $\hat{\mathbf{x}}_{0}^\text{tgt}$:
\begin{align}
    \hat{\mathbf{x}}_{0}^{\text{tgt}} &= \renderer( \text{NeRF}_{\phi}(\mathbf{x}^{\text{tgt}}_t, \mathbf{x}^{\text{c}}, t)).
    \label{eq:one-step-diffusionwithforward} 
\end{align}Similar to~\cref{eq:one-step-mvd}, the \emph{one-step} estimation of $\hat{\mathbf{x}}_{0}^{\text{tgt}}$ can be noisy and inaccurate, especially when $t$ is large. Thus, one can use standard DDPM to sample $\mathbf{x}^{\text{tgt}}_{t-1}$ using~\cref{eq:ddpm_reverse} and perform the iterative denoising.

Inspired by Diffusion-with-Forward~\cite{Tewari2023DiffusionWithForward}, we learn the image-conditional 3D-GS generation in a diffusion-based framework. 
In this scenario, 3D-GS is efficient and thus more appropriate than NeRF for iterative sampling and rendering. 
Our $\renderer(\cdot)$ is the differentiable rasteraizer implemented and accelerated by CUDA, which achieves around 2700 times faster rendering than volume-rendering-based $\renderer(\cdot)$ in~\cite{Tewari2023DiffusionWithForward}.
Moreover, our 3D-GS diffusion model can be enhanced by the 2D multi-view priors from 2D diffusion models in~\cref{sec:mvd}.
We describe this model in more details in~\cref{subsec:MVR}.

\label{sec:3ddifusion}

\section{Gen-3Diffusion}
\label{sec:gen3diffusion}

\begin{figure*}[t!]
  \centering
  \includegraphics[width=\linewidth]{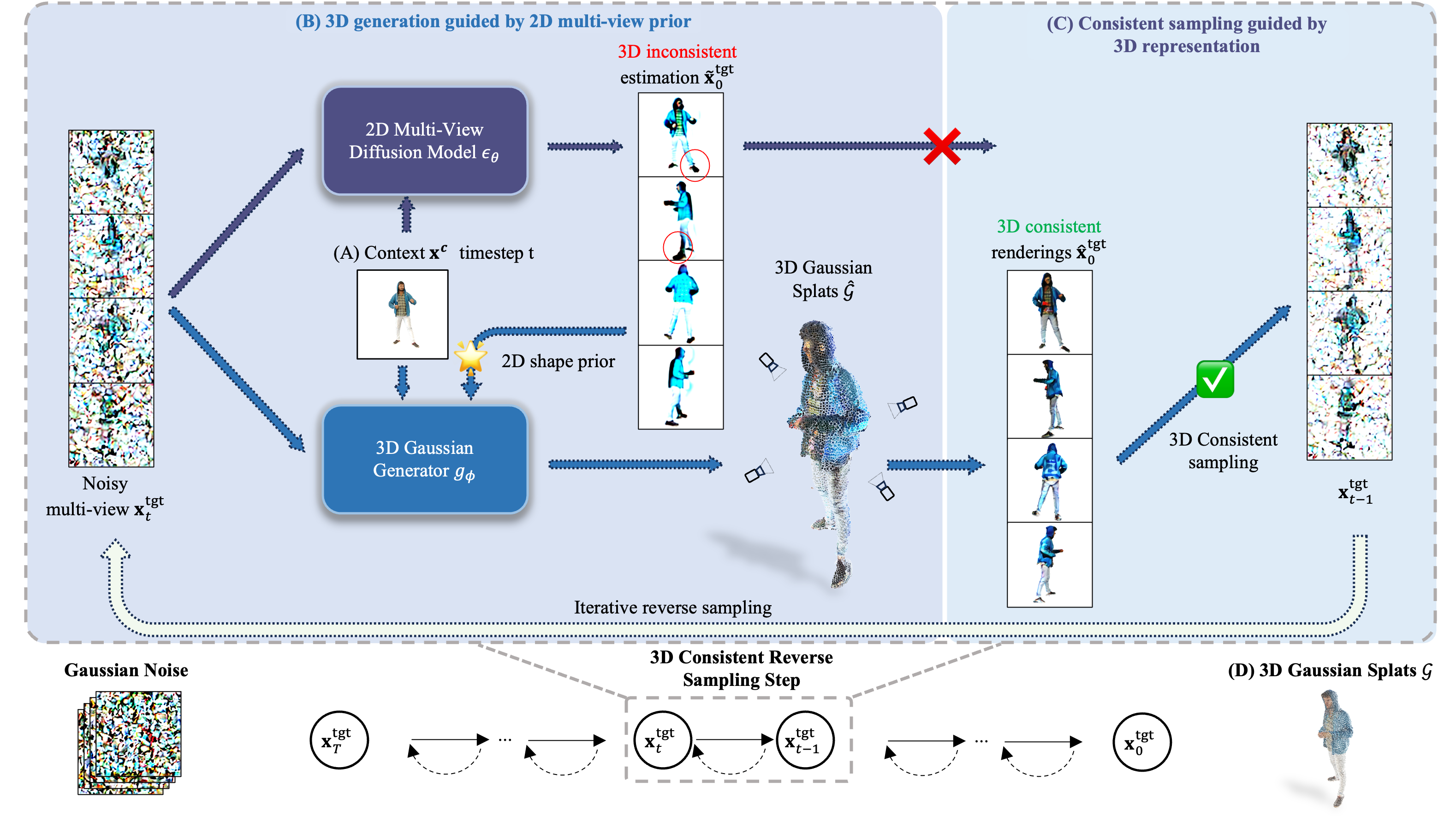}
  \vspace{-0.5cm}
  \caption{\textbf{Method Overview.} Given a single RGB image (A), we sample a realistic 3D object represented as 3D Gaussian Splatting (D) from our learned distribution. At each reverse step, our 3D generation model $g_\phi$ leverages 2D multi-view diffusion prior from $\epsilon_\theta$ which provides a strong shape prior but is not 3D consistent (B,~\cref{sec:2Dhelps3D}). We then refine the 2D reverse sampling trajectory with generated 3D renderings that are guaranteed to be 3D consistent (C,~\cref{sec:3dhelps2d}). Our tight coupling ensures 3D consistency at each sampling step and obtains high-quality 3D Gaussian Splats.    
  }
  \label{fig:pipeline}
\end{figure*}

\noindent\textbf{Overview.} Given a single RGB image, we aim to create a realistic 3D model consistent with the input. We adopt an image-conditioned 3D generation paradigm due to inherent ambiguities in the monocular view. We introduce a novel 3D Gaussian Splatting (3D-GS~\cite{Kerbl20233dgs}) diffusion model that combines shape priors from 2D multi-view diffusion models with the explicit 3D-GS representation. This allows us to jointly train our 3D generative model and a 2D multi-view diffusion model end-to-end and improves the 3D consistency of 2D multi-view generation at inference time. 

In this section, we first introduce our novel generative 3D-GS reconstruction model in~\cref{subsec:MVR}. We then describe how we leverage the 3D reconstruction to generate 3D consistent multi-view results by refining the reverse sampling trajectory (\cref{sec:3dhelps2d}) of 2D diffusion model. An overview of our 2D \& 3D diffusion synergy can be found in \cref{fig:pipeline}.

\subsection{3D-GS Diffusion Model}
\label{subsec:MVR}
Given a context image $\mathbf{x}^c$, we use a conditional diffusion model to learn and sample from a plausible 3D distribution. Previous works demonstrated that 3D generation can be done implicitly via diffusing rendered images of a differentiable 3D represetation~\cite{anciukevicius2023renderdiffusion, karnewar2023holodiffusion, Tewari2023DiffusionWithForward} such as NeRF~\cite{Mildenhall2020NeRF, Yu2021pixelnerf}.

In this work, we propose a 3D-GS diffusion model $g_\phi$, which is conditioned on input context image $\mathbf{x}^c$ to perform reconstruction of 3D Gaussian Splats $\mathcal{G}$. 
Diffusing directly in the space of $\mathcal{G}$ parameters requires pre-computing Gaussian Splats from scans, which is exorbitant. Instead, we diffuse the multi-view renderings of $\mathcal{G}$ using a differentiable rendering function $\renderer(\cdot)$ to learn the conditional 3D distribution. 

We denote $\mathbf{x}^\text{tgt}_0$ as the ground truth images at target views to be diffused and $\mathbf{x}^\text{novel}_0$ as the additional novel views for supervision. 
At training time, we uniformly sample a timestep $t\sim\mathcal{U}(0, T)$ and add noise to $\mathbf{x}^\text{tgt}_0$ using~\cref{eq:ddpm_forward} to obtain noisy target views $\mathbf{x}^\text{tgt}_t$. 
Our generative model $g_\phi$ takes $\mathbf{x}^\text{tgt}_t$, diffusion timestep $t$, and the conditional image $\mathbf{x}^c$ as input, and estimates 3D Gaussians $\hat{\mathcal{G}}$:
\begin{equation}
\begin{gathered}
    \hat{\mathcal{G}} = g_\phi(\mathbf{x}^\text{tgt}_t, t, \mathbf{x}^c), \\ \text{ where } \mathbf{x}^\text{tgt}_t=\sqrt{\Bar{\alpha}_t}\mathbf{x}_0^\text{tgt} + \sqrt{1- \Bar{\alpha}_t}\epsilon, \text{ and } \epsilon\sim\mathcal{N}(0, \mathbf{I})
    \label{eq:diffusion_wo_x0}
\end{gathered}
\end{equation}  

We adopt an asymmetric U-Net Transformer proposed by~\cite{Tang2024LGM} for $g_\phi$ to directly predict 3D-GS parameters from per-pixel features of the last U-Net layer. The context image $\mathbf{x}^c$ is attended onto the noisy image $\mathbf{x}^\text{tgt}_t$ using dense pixel-wise attention.
More specifically, the $\text{H}\times\text{W}\times\text{14}$ feature map is reshaped in $\text{H}*\text{W}\times\text{14}$, where a total number of $\text{H}*\text{W}$ 3D-GS are available, each has a center ${\mathbf{o}\in\mathbb{R}^3}$, a scaling factor ${\mathbf{s}\in\mathbb{R}^3}$, a rotation quaternion ${\mathbf{q}\in\mathbb{R}^4}$, an opacity value ${\boldsymbol{\alpha}\in\mathbb{R}^1}$, and a color feature ${\mathbf{c}\in\mathbb{R}^3}$. For more implementation details regarding the asymmetric U-Net Transformer, please refer to~\cite{Tang2024LGM}.

To supervise the generative model $g_\phi$, we use a differentiable rendering function $\renderer(\cdot): \{\mathcal{G}, \pi^\text{p}\}\mapsto \mathbf{x}^\text{p}$ to render images at target views $\pi^\text{tgt}$ and additional novel views $\pi^\text{novel}$. 
Denoting $\mathbf{x}_0:=\{\mathbf{x}^{\text{tgt}}_{0}, \mathbf{x}^{\text{novel}}_{0}\}$ as ground truth and $\hat{\mathbf{x}}_0:=\{\hat{\mathbf{x}}^{\text{tgt}}_{0}, \hat{\mathbf{x}}^{\text{novel}}_{0}\}$ as rendered images, we compute the loss on images and generated 3D-GS: 
\begin{equation}
    \begin{split}
        \mathcal{L}_{gs} &= \lambda_1 \cdot \mathcal{L}_\text{MSE}\big(\mathbf{x}_0, \hat{\mathbf{x}}_0\big) 
     + \lambda_2 \cdot \mathcal{L}_\text{Percep}\big(\mathbf{x}_0, \hat{\mathbf{x}}_0\big) \\ 
    & + \lambda_3 \cdot \mathcal{L}_\text{reg}(g_\phi(\mathbf{x}^\text{tgt}_t, t, \mathbf{x}^c)), \\
        \text{ where } \hat{\mathbf{x}}_0 &:= \{\hat{\mathbf{x}}^\text{tgt}_0, \hat{\mathbf{x}}^\text{novel}_0\} 
        \\ &= \renderer(g_\phi(\mathbf{x}^\text{tgt}_t, t, \mathbf{x}^c), \{\pi^\text{tgt}, \pi^\text{novel}\}),
    \end{split} 
    \label{eq:loss_3D_diffusion}
\end{equation}
here $\mathcal{L}_\text{MSE}$ denotes the Mean Square Error (MSE) and $\mathcal{L}_\text{Percep}$ is the perceptual loss based on VGG-19~\cite{Simonyan2015vgg}. We also apply $\mathcal{L}_\text{reg}$, a geometry regularizer~\cite{Huang20242dgs, Yu2024gof} to stabilize the generation of $\hat{\mathcal{G}}$.

With this, we can train a generative model that diffuses 3D-GS \emph{implicitly} by diffusing 2D images $\mathbf{x}^\text{tgt}_t$. At inference time, we can generate 3D-GS given the input image by denoising 2D multi-views sampled from Gaussian distribution. We initialize $\mathbf{x}^\text{tgt}_T$ from $\mathcal{N}(0, \mathbf{I})$, and iteratively denoise the rendered images of predicted $\hat{\mathcal{G}}$ from our model $g_\phi$. 
At each reverse step, our model $g_\phi$ estimates a clean state $\hat{\mathcal{G}}$ and render target images $\hat{\mathbf{x}}^\text{tgt}_0$. 
We then calculate target images $\mathbf{x}^\text{tgt}_{t-1}$ for the next step via \cref{eq:xt_minus_one_from_xt_x0_old} and repeat the process until $t=0$, obtaining clear images $\hat{\mathbf{x}}^\text{tgt}_0$ and a clean 3D-GS $\hat{\mathcal{G}}$.

Our generative 3D-GS reconstruction model archives superior performance on in-distribution reconstruction yet generalizes poorly to unseen categories (\cref{subsec:ablation} ~\cref{fig:ablate_multiview_cond_obj_gso}). 
Our key insight for better generalization is leveraging strong priors from pretrained 2D multi-view diffusion models for 3D-GS generation. 

\algrenewcommand\algorithmicrequire{\textbf{Input:}} 
\algrenewcommand\algorithmicensure{\textbf{Output:}}
\algrenewcommand\algorithmicindent{0.5em}%
\begin{figure*}[ht]
\begin{minipage}[t]{0.49\textwidth}
\begin{algorithm}[H]
  \caption{Joint 2D \& 3D Diffusion Training} \label{alg:training}
  \small
  \begin{algorithmic}[1]
  \Require Dataset of posed multi-view images $\mathbf{x}^{\text{tgt}}_{0}$, $\pi^{\text{tgt}}$, $\mathbf{x}^{\text{novel}}_{0}$, $\pi^{\text{novel}}$, a context image $\mathbf{x}^{\text{c}}$, text description $y$
  \Ensure Optimized 2D multi-view diffusion model
  $\epsilon_{\theta}$ and 3D-GS generative model $g_{\phi}$
    \Repeat
      \State $\{\mathbf{x}^{\text{tgt}}_{0}, \mathbf{x}^{\text{novel}}_{0}, \mathbf{x}^{\text{c}}, y\} \sim q(\{\mathbf{x}^{\text{tgt}}_{0}, \mathbf{x}^{\text{novel}}_{0}, \mathbf{x}^{\text{c}}, y\} )$
      \State $t \sim \mathrm{Uniform}(\{1, \dotsc, T\})$; $\boldsymbol{\epsilon} \sim \mathcal{N}(\mathbf{0},\mathbf{I})$
      \State $\mathbf{x}^{\text{tgt}}_{t} = \sqrt{\bar\alpha_t} \mathbf{x}_{0}^{\text{tgt}} + \sqrt{1-\bar{\alpha}_t}\boldsymbol{\epsilon}$
      \State $\tilde{\mathbf{x}}^{\text{tgt}}_0 = \frac{1}{\sqrt{\bar{\alpha}_t}}( \mathbf{x}^{\text{tgt}}_t-\sqrt{1-\bar{\alpha}_t} \boldsymbol{\epsilon}_{\theta}(\mathbf{x}^{\text{tgt}}_t, \mathbf{x}^{\text{c}}, y, t))$ 
      \State $\hat{\mathcal{G}} = g_{\phi}\left(\mathbf{x}_t^{\text{tgt}}, t, \mathbf{x}^{\text{c}}, \tilde{\mathbf{x}}_{0}^{\text{tgt}}\right) $ 
      \textcolor{gray}{// Enhance conditional 3D generation with 2D diffusion prior $\tilde{\mathbf{x}}_{0}^{\text{tgt}}$ from $\mathbf{\epsilon}_\theta$}
      \State $\{ \hat{\mathbf{x}}_{0}^{\text{tgt}}, \hat{\mathbf{x}}_{0}^{\text{novel}}\} = \renderer\left(\hat{\mathcal{G}}, \{\pi^{\text{tgt}}, \pi^{\text{novel}}\} \right)$
      \State Compute loss $\mathcal{L}_{total}$ (~\cref{eq:loss_all}) 
      \State Gradient step to update $\mathbf{\epsilon}_\theta, g_\phi$

    \Until{converged}
  \end{algorithmic}
  \label{algm:train}
\end{algorithm}
\end{minipage}
\hfill
\begin{minipage}[t]{0.49\textwidth}
\begin{algorithm}[H]
  \caption{3D Consistent Guided Sampling} \label{alg:sampling}
  \small
  \begin{algorithmic}[1]
  \Require A context image $\mathbf{x}^c$ and text $y$; Converged 2D diffusion model $\epsilon_{\theta}$ and 3D generative model $g_{\phi}$
  \Ensure 3D Gaussian Splats $\mathcal{G}$ of the 2D image $\mathbf{x}^c$
  
    \vspace{.08in}
    \State $\mathbf{x}_T^{\text{tgt}} \sim \mathcal{N}(\mathbf{0}, \mathbf{I})$
    \For{$t=T, \dotsc, 1$}
      \State $\tilde{\mathbf{x}}_{0}^{\text{tgt}} = \frac{1}{\sqrt{\bar{\alpha}_t}}( \mathbf{x}^{\text{tgt}}_t-\sqrt{1-\bar{\alpha}_t} \boldsymbol{\epsilon}_{\theta}(\mathbf{x}^{\text{tgt}}_t, \mathbf{x}^{\text{c}}, y, t))$ 
      \State  $\hat{\mathcal{G}} = g_{\phi}\left(\mathbf{x}_t^{\text{tgt}}, t, \mathbf{x}^{\text{c}}, \tilde{\mathbf{x}}_{0}^{\text{tgt}}\right) $
      \State $\hat{\mathbf{x}}_{0}^{\text{tgt}} = \renderer\left(\hat{\mathcal{G}}, \pi^{\text{tgt}}\right)$
      \State $\mu_{t-1}(\mathbf{x}_t^\text{tgt}, \hat{\mathbf{x}}_0^\text{tgt}) = \frac{\sqrt{\alpha_{t}}\left(1-\bar{\alpha}_\text{t-1}\right)}{1-\bar{\alpha}_{t}} \mathbf{x}^{\text{tgt}}_{t} + \frac{\sqrt{\bar{\alpha}_\text{t-1}} \beta_{t}}{1-\bar{\alpha}_{t}} \hat{\mathbf{x}}_{0}^{\text{tgt}}$ \textcolor{gray}{// Guide 2D sampling with 3D consistent multi-view renderings}
      \State $\mathbf{x}^{\text{tgt}}_{t-1} \sim \mathcal{N}\left(\mathbf{x}^{\text{tgt}}_{t-1}; \tilde{\boldsymbol{\mu}}_{t}\left(\mathbf{x}^{\text{tgt}}_t, \hat{\mathbf{x}}^{\text{tgt}}_{0} \right), \tilde{\beta}_{t-1}\mathbf{I}) \right)$
    \EndFor
    \vspace{.09in}
    \State \textbf{return} $\mathcal{G} =  g_{\phi}\left(\mathbf{x}_{0}^{\text{tgt}}, \tilde{\mathbf{x}}_{0}^{\text{tgt}}, \mathbf{x}^{\text{c}}, t=0\right) $
     \vspace{.015in}
  \end{algorithmic}
\end{algorithm}
\label{algm:sample}
\end{minipage}
\vspace{-1em}
\end{figure*}

\subsection{3D Diffusion with 2D Multi-View Priors}
\label{sec:2Dhelps3D}
Pretrained 2D multi-view diffusion models (MVD)~\cite{liu2023syncdreamer, shi2023mvdream, Wang2023ImageDream} have seen billions of real images~\cite{Schumann2022Laion5B} and millions of 3D data~\cite{Deitke2023Objaverse}, which provide strong prior information and can generalize to unseen objects~\cite{Tang2024LGM, xu2024instantmesh}. Here, we propose a simple yet elegant idea for incorporating this multi-view prior into our generative 3D-GS model $g_\phi$. We can also leverage generated 3D-GS to guide 2D MVD sampling process which we discuss in~\cref{sec:3dhelps2d}.\\
Our key observation is that both 2D MVD and our proposed 3D-GS generative model are diffusion-based and share the same sampling state $\mathbf{x}_{t}^{\text{tgt}}$ at timestep $t$. Thus, they can be tightly \emph{synchronized}. This enables us to couple and facilitate information exchange between 2D MVD $\boldsymbol{\epsilon}_{\theta}$ and 3D-GS generative model $g_{\phi}$ at the same diffusion timestep $t$.  
To inject the 2D diffusion priors into 3D generation, we first compute \emph{one-step} estimation of $\tilde{\mathbf{x}}^{\text{tgt}}_0$ (\cref{eq:one-step-mvd}) using 2D MVD $\epsilon_\theta$, and condition our 3D-GS generative mode $g_{\phi}$ additionally on it. Formally, our 3D-GS generative model enhanced with 2D multi-view diffusion priors is written as:
\begin{equation}
\begin{gathered}
    \hat{\mathcal{G}} = g_\phi(\mathbf{x}^\text{tgt}_t, t, \mathbf{x}^c, \Tilde{\mathbf{x}}^\text{tgt}_0), \\\text{ where } \tilde{\mathbf{x}}_{0}^{\text{tgt}} = \frac{1}{\sqrt{\bar{\alpha}_t}}( \mathbf{x}^{\text{tgt}}_t-\sqrt{1-\bar{\alpha}_t} \boldsymbol{\epsilon}_{\theta}(\mathbf{x}^{\text{tgt}}_t, \mathbf{x}^{\text{c}}, t)).
    \label{eq:diffusion_with_x0_clean}
\end{gathered}
\end{equation}

\begin{figure*}[t!]
  \centering
  \includegraphics[width=\textwidth]{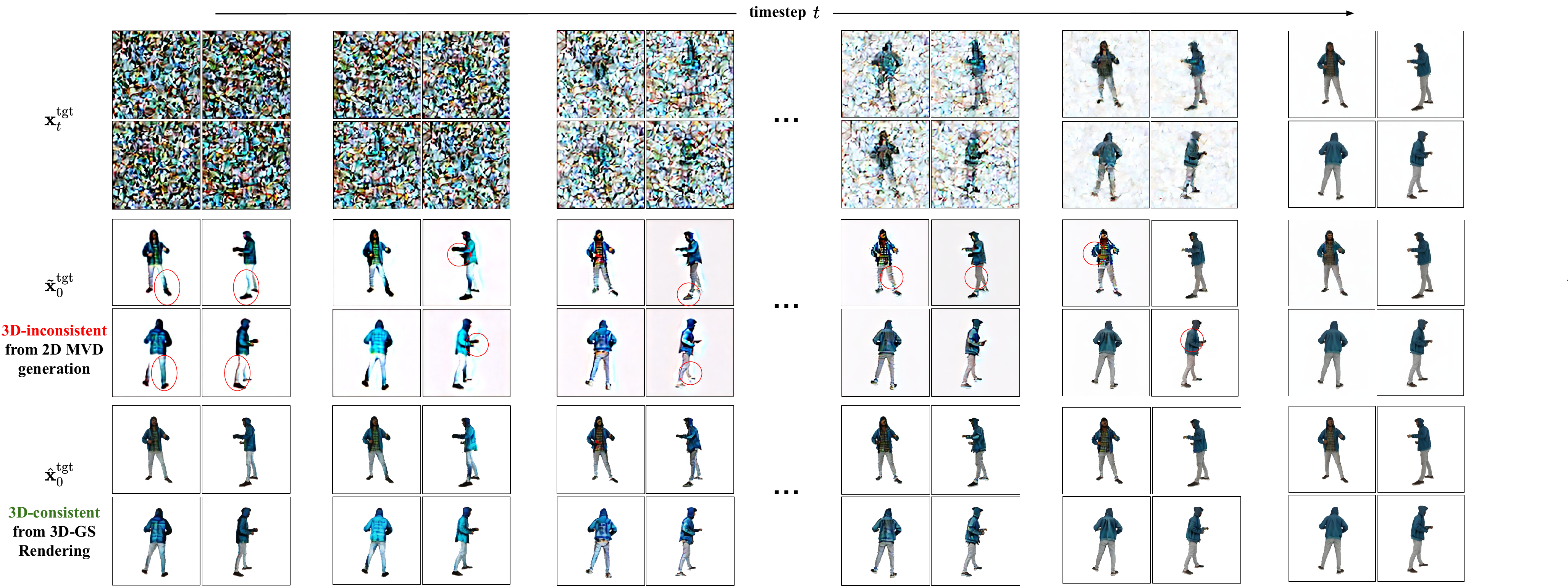}
  \vspace{-0.5cm}
  \caption{Visualization of DDPM reverse sampling trajectory. At each individual step, estimated $\tilde{\textbf{x}}_{0}^{\text{tgt}}$ can be 3D inconsistency across different views, while the rendering $\hat{\textbf{x}}_{0}^{\text{tgt}}$ are 3D consistent and can refine the inconsistency along trajectory (~\cref{eq:xt_minus_one_from_xt_x0_ours}).}
  \label{fig:ddim_intermediate_visualization}
\end{figure*}

\begin{figure*}[t!]
    \begin{minipage}[t]{0.4\textwidth} 
    \footnotesize
    \strut\vspace*{-0.6cm}\newline
        \centering
        \captionof{table}{Ablation of 2D multi-view priors in o.o.d. generalization.}
         \begin{tabular}{ l c c c c }
            \hline
            Method & { LPIPS$\downarrow$} & {SSIM$\uparrow$}  &{PSNR$\uparrow$}   & {FID$\downarrow$} \\
             \hline
            Ours w/o $ \tilde{\mathbf{x}}^{\text{tgt}}_{0} $ & $0.189$  & $0.721$  & $14.45$  & $107.14$ \\
            \hline
              \emph{Ours} & $\mathbf{0.194}$  & $\mathbf{0.778}$  & $\mathbf{16.12}$  & $\mathbf{83.89}$  \\
             \hline
        \end{tabular}
    \end{minipage}
    \hfill
  \begin{minipage}[t]{0.6\textwidth}
  \strut\vspace*{-\baselineskip}\newline 
    \centering
      \includegraphics[width=\linewidth]{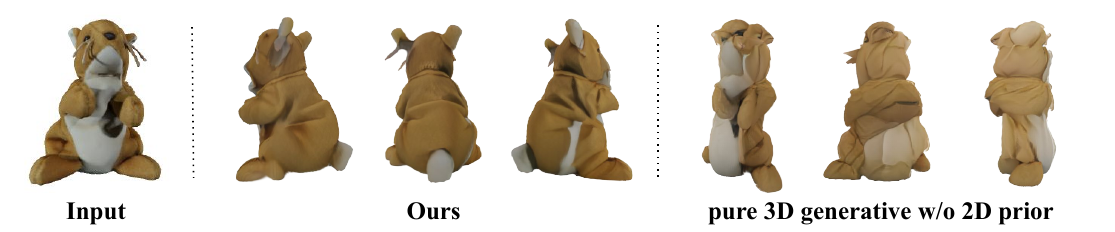}

  \end{minipage}
    \caption{\textbf{Ablation of 2D multi-view priors} for object reconstruction. The model is trained for human reconstruction. It can be seen in the left table that our 2D MVD model improves the generalization ability to unseen objects, leading to more plausible object shape as shown on the right image. 
    }
    \label{fig:ablate_multiview_cond_obj_gso}
\end{figure*}

The visualization of $\tilde{\mathbf{x}}^{\text{tgt}}_0$ along the whole sampling trajectory in~\cref{fig:ddim_intermediate_visualization} shows that the pretrained 2D diffusion model $\boldsymbol{\epsilon}_{\theta}$ can already provide useful multi-view shape prior even in large timestep $t=1000$. This is further validated in our experiments where the additional 2D diffusion prior $\tilde{\mathbf{x}}^{\text{tgt}}_0$ leads to better 3D reconstruction (\cref{tab:ablate_multiview_cond_human}) as well as more robust generalization to general objects (\cref{fig:ablate_multiview_cond_obj_gso}). By utilizing the timewise iterative manner of 2D and 3D diffusion models, we can not only leverage 2D priors for 3D-GS generation but also train both models jointly end to end, which we discuss in~\cref{sec:3dhelps2d}. 

\subsection{Synergy between 2D \& 3D Diffusion}
\label{sec:3dhelps2d}

\noindent\textbf{Joint Diffusion Training} We adopt pretrained ImageDream~\cite{Wang2023ImageDream} as our 2D multi-view diffusion model $\epsilon_\theta$ and jointly train it with our 3D-GS generative model $g_\phi$. We observe that our joint training is important for coherent 3D generation, as opposed to prior works that frozen pretrained 2D multi-view models~\cite{Tang2024LGM, tochilkin2024triposr}. 
We summarize our training algorithm in~\cref{alg:training}. 
We combine the loss of 2D diffusion and our 3D-GS generation loss $\mathcal{L}_{gs}$(~\cref{eq:loss_3D_diffusion}): 
\begin{equation}
    \begin{split}
    \mathcal{L}_{total} = \mathcal{L}_\text{MSE}(\boldsymbol{\epsilon}, \boldsymbol{\epsilon}_{\theta}) + \mathcal{L}_{gs}
    \end{split}
    \label{eq:loss_all}
\end{equation}
Once trained, one can sample a plausible 3D-GS $\mathcal{G}$ conditioned on the input image from the learned 3D distributions. 
However, we observe that the multi-view diffusion model $\boldsymbol{\epsilon}_{\theta}$ can still output inconsistent multi-views along the sampling trajectory (see~\cref{fig:pipeline}). On the other hand, our 3D generator produces explicit 3D-GS which can be rendered as 3D consistent multi-views. Our second key idea is to use the 3D consistent renderings to guide 2D sampling process for more 3D consistent multi-view generation. We discuss this next.

\noindent\textbf{3D Consistent Guided Sampling}
With the shared and \emph{synchronized} sampling state $\mathbf{x}_t^\text{tgt}$ of 2D multi-view diffusion model $\boldsymbol{\epsilon}_{\theta}$ and 3D-GS reconstruction model $g_\phi$, we couple both models at arbitrary $t$ during training. Similarly, they are also connected by both using estimated clean multi-views $\mathbf{x}_0^\text{tgt}$ at sampling time. To leverage the full potential of both models, we carefully design a joint sampling process that utilizes the reconstructed 3D-GS $\hat{\mathcal{G}}$ at each timestep $t$ to guide 2D multi-view sampling, which is summarized in~\cref{alg:sampling}. \\
We observe that the key difference between the clean multi-views estimated $\mathbf{x}_0^\text{tgt}$ from 2D diffusion model and our 3D-GS generation lies in 3D consistency: 2D MVD computes multi-view $\tilde{\mathbf{x}}_0^\text{tgt}$ from 2D network prediction which can be 3D inconsistent while our $\hat{\mathbf{x}}_0^\text{tgt}$ are rendered from explicit 3D-GS representation which are guaranteed to be 3D consistent. Our idea is to guide the 2D multi-view reverse sampling process with our 3D consistent renderings $\hat{\mathbf{x}}_0^\text{tgt}$ such that the 2D sampling trajectory is more 3D consistent. Specifically, we leverage 3D consistent multi-view renderings $\hat{\mathbf{x}}_0^\text{tgt}$ to refine the posterior mean $\mu_\theta(\mathbf{x}_t^\text{tgt}, t)$ at each reverse step: 
\begin{align}
\begin{split}
    \text{Original: } &\mu_\theta(\mathbf{x}_t^\text{tgt}, t) :=\mu_{t-1}(\mathbf{x}_t^\text{tgt}, \tilde{\mathbf{x}}_0^\text{tgt}) \quad \\ \rightarrow \quad \text{Ours: } &\mu_\theta(\mathbf{x}_t^\text{tgt}, t) :=\mu_{t-1}(\mathbf{x}_t^\text{tgt}, \hat{\mathbf{x}}_0^\text{tgt}),   \\
    \text{ where } \hat{\mathbf{x}}_{0}^{\text{tgt}}&=\renderer(\hat{\mathcal{G}}, \pi^\text{tgt}), \\
    \mu_{t-1}(\mathbf{x}_t^\text{tgt}, \hat{\mathbf{x}}_0^\text{tgt}) &= \frac{\sqrt{\alpha_{t}}\left(1-\bar{\alpha}_{t-1}\right)}{1-\bar{\alpha}_{t}} \mathbf{x}^{\text{tgt}}_{t} + \frac{\sqrt{\bar{\alpha}_{t-1}} \beta_{t}}{1-\bar{\alpha}_{t}}  \hat{\mathbf{x}}_{0}^{\text{tgt}} 
\end{split}
    \label{eq:xt_minus_one_from_xt_x0_ours}
\end{align}

With this refinement, we guarantee the 3D consistency at each reverse step $t$ and avoid 3D inconsistency accumulation in original multi-view sampling~\cite{Wang2023ImageDream}. 
In~\cref{fig:ddim_intermediate_visualization}, we visualize the evolution of originally generated multi-views $\tilde{\mathbf{x}}^{\text{tgt}}_0$ and multi-views rendering $\hat{\mathbf{x}}^{\text{tgt}}_0$ from generated 3D-GS $\hat{\mathcal{G}}$ along the whole reverse sampling process. It intuitively shows how effective the sampling trajectory refinement is. We perform extensive ablation in \cref{subsec:ablation}, which shows the importance of consistent refinement for the sampling trajectory. 
\section{Experiments}
\label{sec:experiments}
In this section, we demonstrate the effectiveness of our method for two image-based 3D reconstruction tasks: general object reconstruction (denoted as $\textit{Gen3D}_\text{object}$, \cref{subsec:exp-object}) and clothed human avatar reconstruction (denoted as $\textit{Gen3D}_\text{avatar}$, \cref{subsec:exp-human}). We also ablate the influence of 2D and 3D diffusion models to our full pipeline in \cref{subsec:ablation}. 

\subsection{Experimental Setup}
\begin{figure*}[htbp]
    \centering
    \begin{minipage}[b]{0.49\textwidth}
        \centering
        \includegraphics[width=\textwidth]{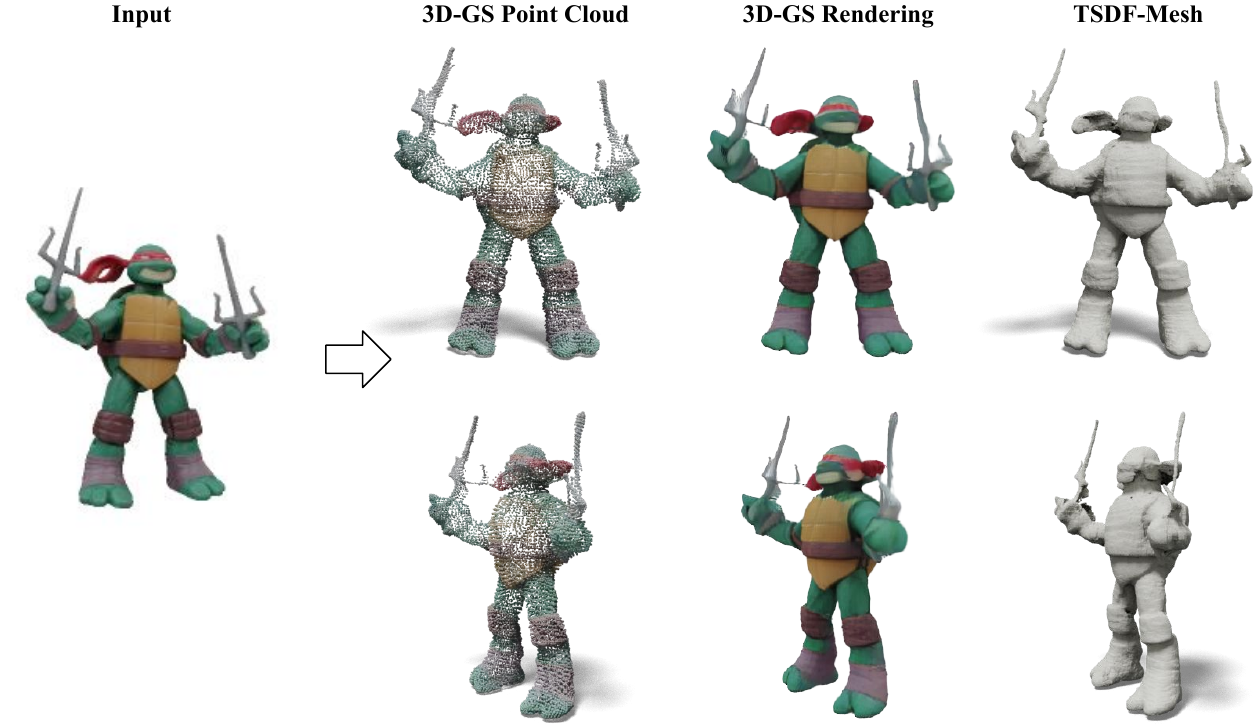} 
        \vspace{-0.5cm}
        \caption{Visualization of outputs in our pipeline. From left to right: (a) input RGB image; (b) reconstructed 3D Gaussian Splats point cloud; (c) novel-view renderings produced by splatting the 3D-GS; and (d) high-fidelity textured mesh extracted via TSDF fusion of multi-view depth maps.
        }
        \label{fig:visualization_modality}
    \end{minipage}
    \hfill
    \begin{minipage}[b]{0.49\textwidth}
        \centering
        \includegraphics[width=\textwidth]{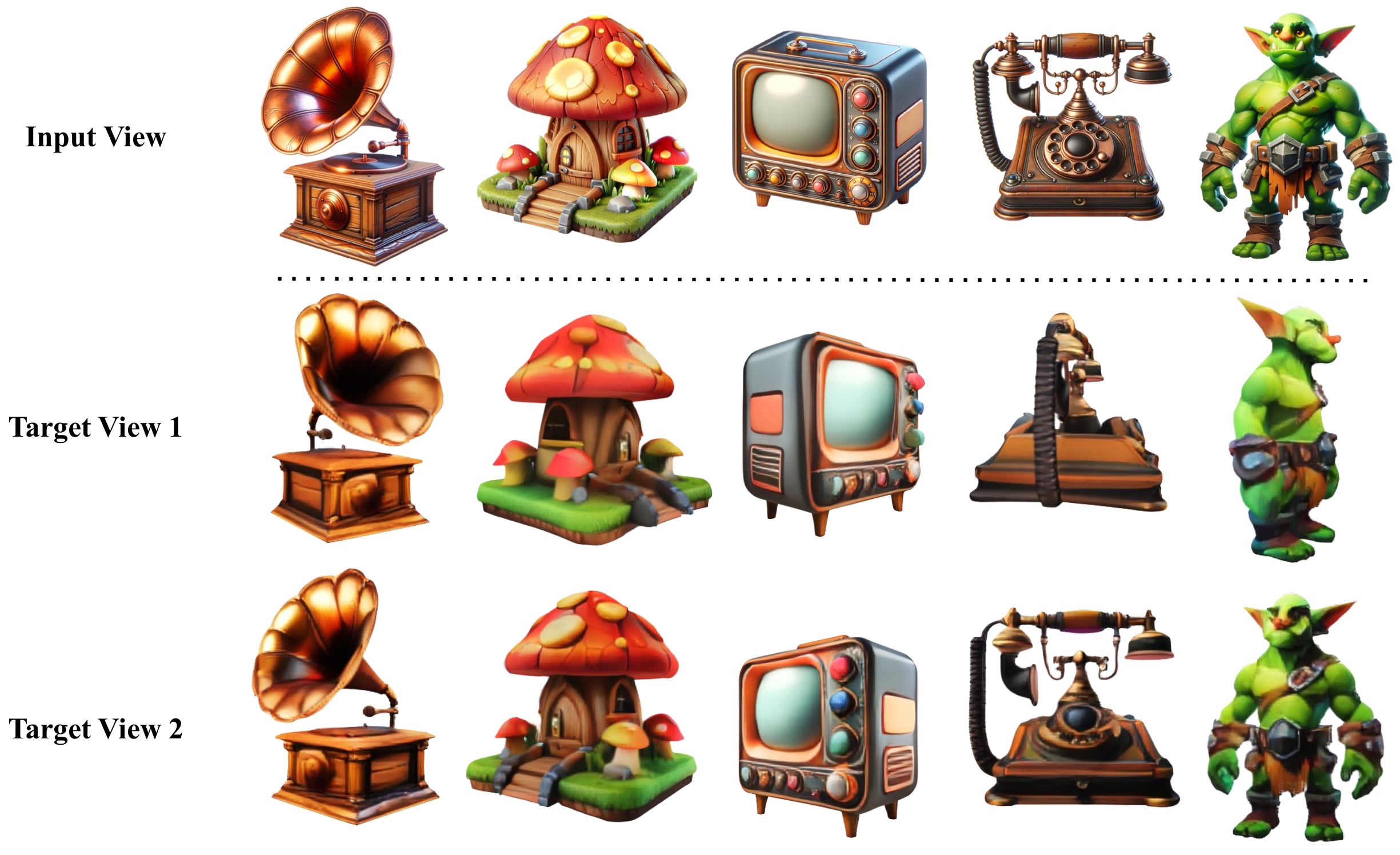} 
        \vspace{-0.5cm}
        \caption{Qualitative results on in-the-wild objects. The top row shows the input view for each test object, while the second and third rows display two automatically generated novel views, demonstrating our method’s ability to generalize to diverse object and input camera poses.
        }
        \label{fig:art_alone}
    \end{minipage}
\end{figure*}
\subsubsection{Datasets} 
\noindent\textbf{General Object.}
We use a filtered high-quality Objaverse~\cite{Deitke2023Objaverse} subset introduced in LGM~\cite{Tang2024LGM} which consists of around 80K objects to train our $\textit{Gen3D}_\text{object}$ model. We evaluate our model on the Google Scanned Object dataset (GSO)~\cite{downs2022gso} according to the same evaluation protocol specified in EscherNet~\cite{kong2024eschernet}.

\noindent\textbf{Clothed Avatar.} We train our $\textit{Gen3D}_\text{avatar}$ model on a combined 3D human dataset~\cite{axyz, treedy, twindom, renderpeople, han20232k2k, ho2023customhuman, su2022thuman3, tao2021thuman2}, compromising  $\sim6000$ high quality scans. 
We evaluate our $\textit{Gen3D}_\text{avatar}$ on around 450 subjects from three different datasets: sizer~\cite{antic2024closed, tiwari2020sizer}, iiit~\cite{jinka2023iiit}, and cape~\cite{ma2019cape}. 

\subsubsection{Implementation Details}
\noindent\textbf{Camera Poses Definition.}
For 2D multi-view diffusion, we adopted the pre-trained model from ImageDream~\cite{Wang2023ImageDream} and slightly modify the camera poses from absolute coordinate system to relative coordinate system. We generate 4 orthogonal views, which has 0, 90, 180, 270 \textit{relative} azimuth and 0 \textit{absolute} elevation given an input view.

\noindent\textbf{Network Architecture.}
Following~\cite{Tang2024LGM}, our 3D-GS generative model $g_{\phi}$ consists of 6 down blocks, 1 middle block, and 5 up blocks, with the input image at $256\times256$ and output Gaussian feature map at 128 $\times$ 128. For each iteration, we start from 4 Gaussian noisy images $\mathbf{x}^{\text{tgt}}_t$ and concatenating their corresponding 2D prior images $\tilde{\mathbf{x}}^{\text{tgt}}_0$ channel-wise, and our $g_{\phi}$ generates in total $128\times128\times4=65536$ number of 3D-GS. For implementation details regarding the U-Net model, please refer to~\cite{Tang2024LGM, Rombach2022StableDiffusion}.

\noindent\textbf{Training.}
We trained both our $\textit{Gen3D}_\text{object}$ and $\textit{Gen3D}_\text{avatar}$ models on 8 NVIDIA A100 GPUs for approximately 5 days. Each GPU was configured with a batch size 2 and gradient accumulations of 16 steps to achieve an effective batch size of 256. Each batch involved sampling 4 orthogonal images with zero elevation angle as target views $\mathbf{x}^{\text{tgt}}_0$, and 12 additional images as novel views $\mathbf{x}^{\text{novel}}_0$ to supervise the 3D generative model~\cref{eq:loss_all}. The hyperparameters for training~\cref{eq:loss_all} were set as follows: $\lambda_1=1.0$, $\lambda_2=1.0$, and $\lambda_3=100.0$. During training, we employed the standard DDPM scheduler~\cite{Ho2020DDPM} to construct noisy target images $\mathbf{x}_{t}^{\text{tgt}}$. The maximum diffusion step $T$ is set to $1000$. The AdamW~\cite{kingma14adam} optimizer is adopted with the learning rate of $5 \times 10^{-4}$ , weight decay of 0.05, and betas of (0.9, 0.95). The learning rate is cosine annealed to 0 during the training. We clip the gradient with a maximum norm of 1.
\vspace{0.2cm}
\noindent\textbf{Inference.} Our whole pipeline, including joint sampling of both 2D \& 3D diffusion models, takes only about 11.7 GB of GPU memory and 22.6 seconds for inference on NVIDIA A100, which is friendly for deployment. 
For the adopted pretrained multi-view diffusion model, we use a guidance scale of 5 following~\cite{Wang2023ImageDream}.
At inference time, we use DDIM scheduler~\cite{song2021ddim} to perform faster reverse sampling. The total reverse sampling steps are set to 50 in all experiments.
The generated 3D Gaussian Splats can be directly splatted from any viewpoint to produce novel-view renderings. For mesh extraction, one can use our implementation to first render multi-view depth maps from the 3D-GS using GOF~\cite{Yu2024gof}, then fuse these depth maps with a TSDF volume~\cite{zeng2016tsdf} to recover a textured mesh. A side-by-side visualization of the raw 3D-GS geometry, the corresponding RGB renderings, and the final extracted mesh is shown in~\cref{fig:visualization_modality}.

\subsubsection{Evaluation Metrics} 
We evaluate the 3D reconstruction quality in terms of appearance and geometry. For appearance quality, we compute metrics on directly generated images (novel view synthesis methods) or renderings (direct 3D reconstruction methods) at 32 novel camera views with uniform azimuth and zero elevation angle. 
The metrics for appearance reported include multi-scale Structure Similarity (SSIM)~\cite{wang2003ssim}, Learned Perceptual Image Patch Similarity (LPIPS)~\cite{zhang2018lpips}, and Peak Signal to Noise Ratio (PSNR) between predicted and ground-truth views. Moreover, we report the Fr\'echet inception distance (FID)~\cite{heusel2017fid} between synthesized views and ground truth renderings, which reflects the quality and realism of the unseen regions. 
We also conduct user studies to evaluate the generation quality with human preference. 

For geometry quality, we compute Chamfer Distance ($\text{CD}$), Point-to-Surface distance (P2S), F-score~\cite{Tatarchenko2019fscore} (w/ threshold of $0.01m$), and Normal Consistency (NC) between the extracted geometry and the groundtruth scan. We normalize the extracted geometry into $[-1, 1]$ and perform iterative closest point (ICP) to match the global pose between extracted and groundtruth geometry to ensure alignment, same as ~\cite{ho2023sith, kong2024eschernet}.
In all experiments, we re-evaluate the baseline models by using their officially open-sourced checkpoints on the
same set of reference views for a fair comparison.

\subsection{3D Object from Image}\label{subsec:exp-object}

\begin{figure*}[t!]
  \centering
  \includegraphics[width=\textwidth]{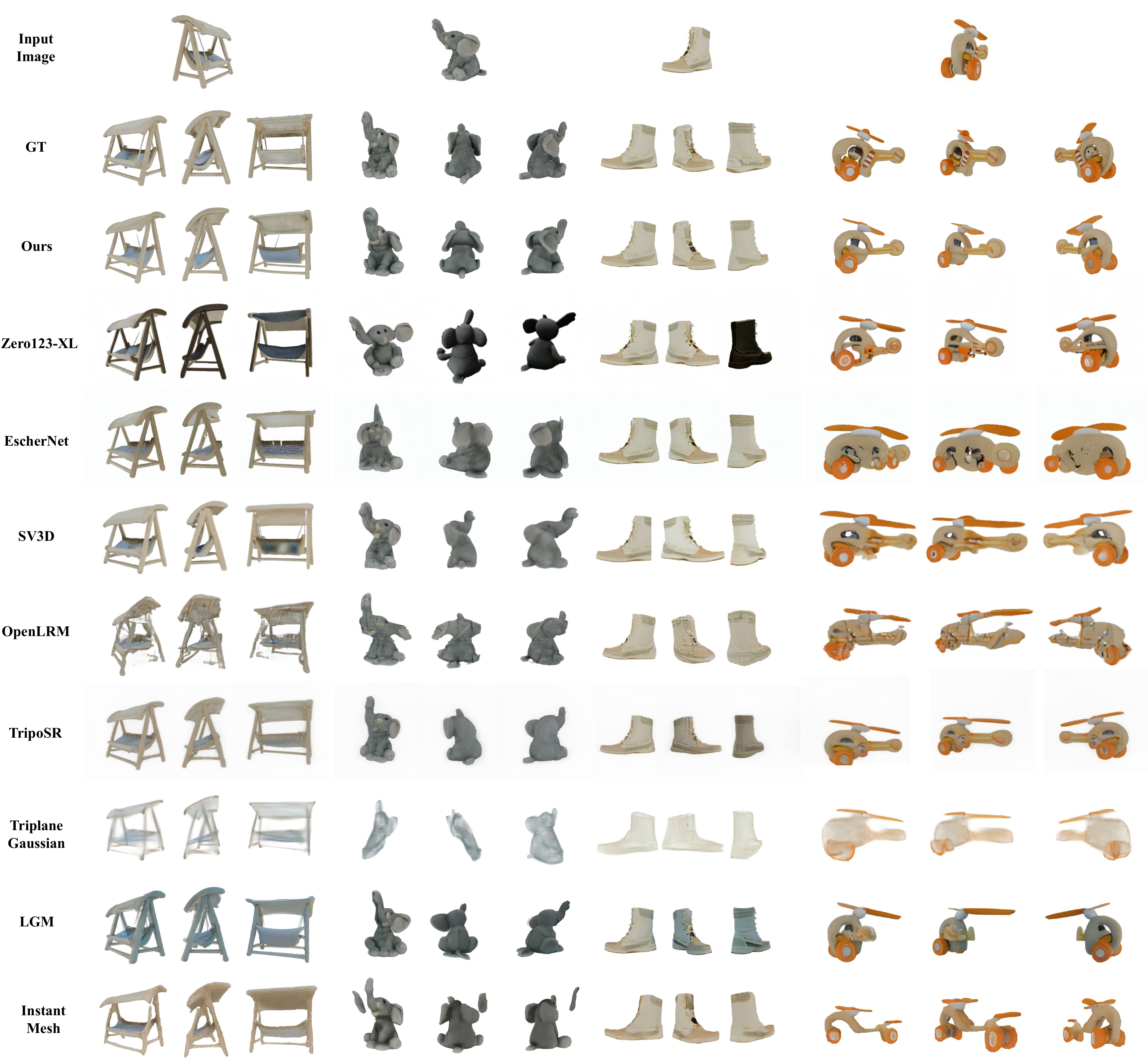}
  \vspace{-0.5cm}
  \caption{\textbf{Novel view synthesis visualization of 3D objects from images.} Our $\textit{Gen3D}_\text{object}$ is able to directly generate 3D-GS and render to arbitrary desired novel multi-views, which are more detailed and faithful w.r.t. the context image, and more 3D-consistent compared to prior works.}
  \label{fig:object_comparison}
\end{figure*}

\begin{table*}[t!]
  \centering
  \caption{\textbf{Comparing object reconstruction methods on GSO dataset~\cite{downs2022gso}}. Our method achieves a better appearance and higher-quality 3D geometry.}
\label{tab:compare_baselines_object}
\begin{tabular}{ l c c c c c c c c}
 \hline
Method  &{PSNR $\uparrow$}  & {SSIM $\uparrow$} & { LPIPS $\downarrow$} &{FID $\downarrow$} & {$\text{CD}_\text{(cm)}$ $\downarrow$}& {$\text{P2S}_\text{(cm)}$ $\downarrow$} &  $\text{NC} \uparrow$ & {F-score $\uparrow$}    \\
     
 \hline
 Zero-1-to-3-XL~\cite{Liu2023zero123} & $ 20.324 $ & $ 0.884 $ & $ 0.107 $  &$ 65.14 $  &  $-$    &  $ - $   & $ - $ & $ - $  \\
 EscherNet~\cite{kong2024eschernet} & $20.503$ & $0.895$ & $0.107$  &$ 65.75 $  & $ - $  &  $-$    & $ - $  &$ - $ \\
 SV3D~\cite{voletiSV3DNovelMultiview2024} & $20.975$ & $0.900$ & $0.105$  &$ 64.72 $  & $ - $   & $ - $ &  $-$    &$ - $ \\
\hline
 OpenLRM~\cite{openlrm} & $ 18.972 $ & $0.880$ & $0.133$  &$ 143.29 $   &  $9.17$  &  $ 9.37 $    & $0.663$  & $0.112$ \\
TripoSR~\cite{tochilkin2024triposr}  & $ 19.820 $ & $ 0.898 $ & $ 0.110 $  &$ 73.26 $  &  $ 6.23 $  &  $ 6.49 $    & $ 0.734 $ &  $0.178$  \\
TriplaneGaussian~\cite{zou2023triplanegaussian} & $18.067 $ & $ 0.893 $ & $0.132$   &$ 149.92 $   &  $ 10.83 $   &  $ 14.18$   & $ 0.601 $  &   $0.081$ \\
LGM  & $19.089$ & $0.885$ & $0.122$  &$ 64.16 $  & $9.88$   &  $ 12.32$   & $0.579$ & $0.146$  \\
InstantMesh  & $18.924 $ & $ 0.886 $ & $0.122$ & $ 101.01 $   &  $ 13.51 $   & $ 15.92$   & $ 0.636 $  &  $0.082$ \\
\hline
$\textit{Gen3D}_\text{object}$ & $\textbf{22.881}$ & $\textbf{0.917}$ & $\textbf{0.078}$  &$ \textbf{54.12} $ &  $\textbf{4.12}$  &  $ \textbf{4.00} $ & $\textbf{0.734}$  &  $\textbf{0.293}$ \\
 \hline
\end{tabular}
\end{table*}

\begin{figure*}[ht]
    \centering
    \begin{minipage}{0.4\textwidth}
        \centering
        \captionof{table}{\textbf{Runtime performance comparison on objects}. We evaluate runtime for generating 32 novel views.}
        \label{tab:inference_efficiency_obj}
        \footnotesize
        \begin{tabular}{ l c c }
        \hline
         &   Time (s) & VRAM (GB)  \\     
         \hline
        Zero-1-to-3~\cite{Liu2023zero123} &  $26.5$ & $3.2$   \\
        EscherNet~\cite{kong2024eschernet} &  $31.4$ & $3.9$ \\
        SV3D~\cite{voletiSV3DNovelMultiview2024} & $109.8$ & $21.5$ \\
        \hline
        OpenLRM~\cite{openlrm} &  $5.9$ & $7.7$ \\
        TripoSR~\cite{tochilkin2024triposr} &  $80.6$ & $5.7$ \\
        TriplaneGaussian~\cite{zou2023triplanegaussian} &  $1.6$ & $4.3$ \\
        LGM~\cite{Tang2024LGM} &  $1.8$ & $7.1$ \\
        \hline
        $\textit{Gen3D}_\text{obj}$ & $23.6$ & $11.7$  \\
         \hline
        \end{tabular}
    \end{minipage}%
    \hfill
    \begin{minipage}{0.6\textwidth}
        \centering
        \includegraphics[width=\textwidth]{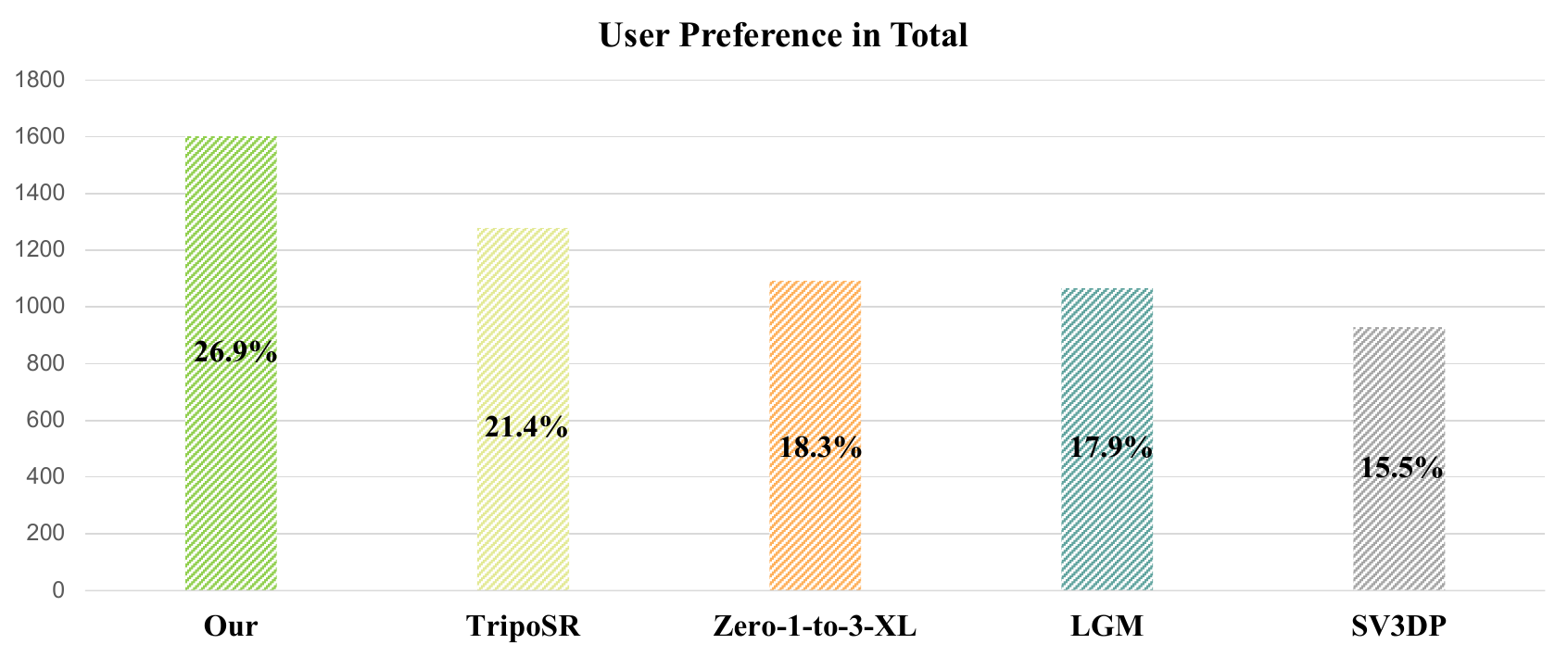}  
    \end{minipage}%
    \caption{\textbf{Runtime performance \& user preference comparison for objects reconstruction.} Left: Inference time and GPU consumption of SoTA object reconstruction approaches. Right: User study statistics of object reconstruction on 360° views. Our method is preferred more than baselines.}
    \label{fig:user-study-obj}
\end{figure*}

\begin{figure*}[t!]
  \centering
  \includegraphics[width=\textwidth]{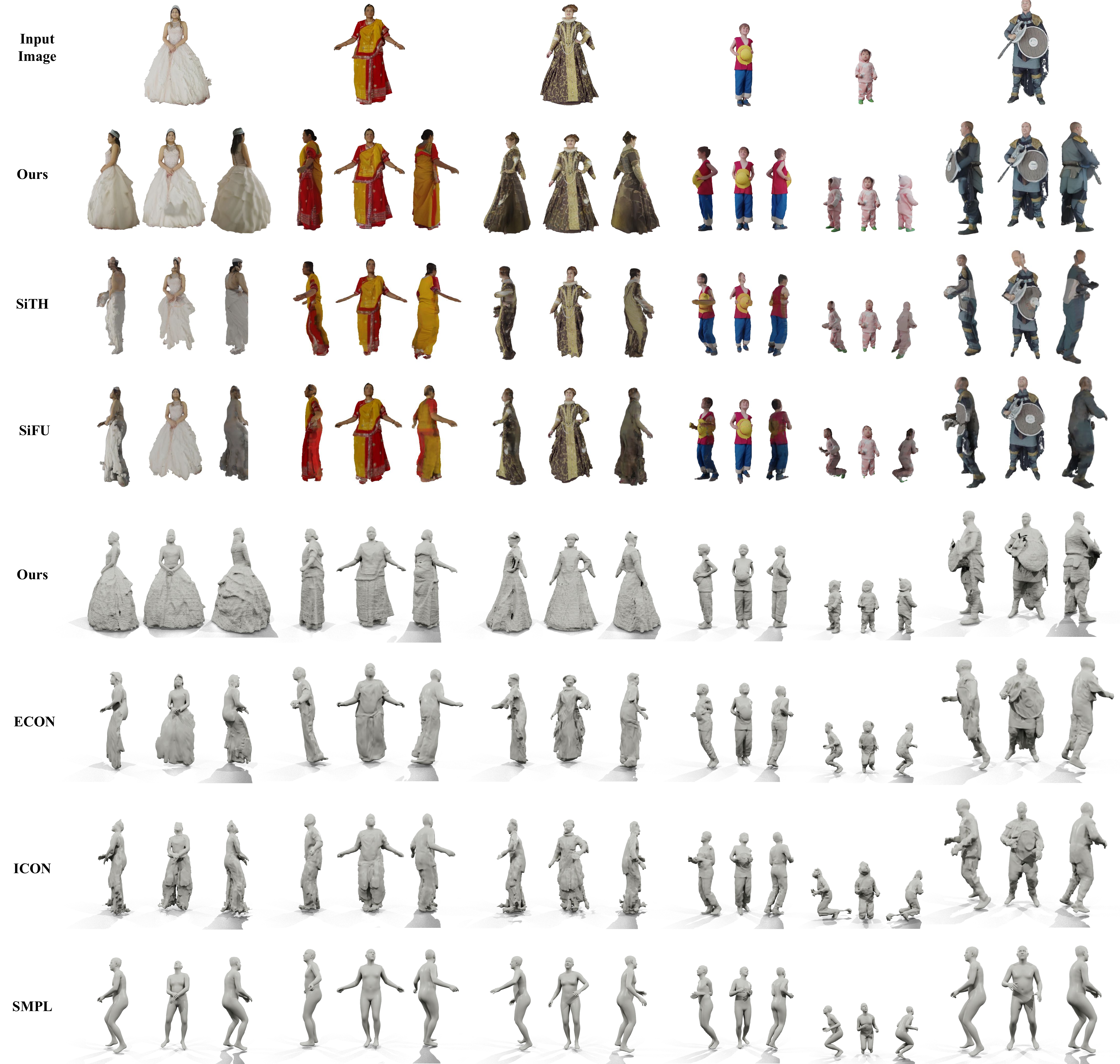}
  \vspace{-0.5cm}
  \caption{\textbf{Appearance and geometry comparison of human avatar creation methods.} We show novel view renderings of the textured avatar and extracted mesh in row 2-4 and row 5-7 respectively. Prior methods produce blurry textures (SiTH~\cite{ho2023sith}, SiFU~\cite{zhang2023sifu}) or oversmoothed surface (ECON~\cite{xiu2023econ}, ICON~\cite{xiu2023icon}) on unseen backside regions. The reliance on SMPL estimation is susceptible to errors (shown in row 8) and makes them difficult to model loose clothing or diverse human shapes. In contrast, our method adopts template-free 3D-GS representation, and leverages strong prior from 2D MVD models, allowing us to faithfully reconstruct high-fidelity geometry and appearance from single RGB image. 
  }
  \label{fig:human_comparison}
\end{figure*}

We evaluate $\textit{Gen3D}_\text{object}$ for novel view synthesis and geometry reconstruction on the GSO dataset~\cite{downs2022gso}. We compare our method against 2D novel view diffusion models Zero-1-to-3~\cite{Liu2023zero123}, Zero-1-to-3-XL~\cite{Liu2023zero123}, EscherNet~\cite{kong2024eschernet}, and SV3D~\cite{voletiSV3DNovelMultiview2024}, as well as methods that directly reconstruct 3D models such as LRM~\cite{hong2023lrm}, TriplaneGaussian~\cite{zou2023triplanegaussian} and LGM~\cite{Tang2024LGM}. 
Since the code for LRM is not publicly available, we adopt the implementation and pretrained model of OpenLRM~\cite{openlrm} and TripoSR~\cite{tochilkin2024triposr} and compare with them.
Notably, many other 2D multi-view diffusion models~\cite{shi2023zero123++, long2023wonder3d, Wang2023ImageDream} prioritize 3D generation rather than view synthesis. This limits their methods to generate fixed target views rather than arbitrary free-view synthesis, making them not directly comparable. 

We report the quantitative evaluation results in \cref{tab:compare_baselines_object} and show some comparisons in \cref{fig:object_comparison}. Novel View Diffusion models~\cite{Liu2023zero123, kong2024eschernet, voletiSV3DNovelMultiview2024} achieve good appearance metrics yet they cannot directly produce a 3D representation from the images. Direct reconstruction approaches~\cite{zou2023triplanegaussian, openlrm, tochilkin2024triposr} predicts 3D directly from images. However, the geometry could be over-smooth (TriplaneGaussian~\cite{zou2023triplanegaussian}) or the texture is not realistic (LGM~\cite{Tang2024LGM}). Our $\textit{Gen3D}_\text{object}$ diffuses 2D images and 3D-GS jointly, which results in better and more 3D-consistent view synthesis and better 3D reconstruction. We also ran a user study with 60 participants (see \cref{fig:user-study-obj}) on the evaluated objects, in which our $\textit{Gen3D}_\text{object}$ was chosen as the most plausible reconstruction for $26.9\%$ of queries, namely 5.5 points higher than the next-best method ($21.4\%$).

We showcase our method on in-the-wild artwork images which are generated by GPT-4o and DALLE 3 in \cref{fig:art_alone}. These results demonstrate that $\textit{Gen3D}_\text{object}$ generalizes beyond real-world photos, producing consistent multi-view renderings even when the input comes from highly varied camera angles and artistic styles.





\subsection{Realistic Avatar from Image}\label{subsec:exp-human}

\begin{table*}[t!]
  \centering
  \caption{\textbf{Geometry evaluation for clothed avatars reconstruction} on Sizer, IIIT, and CAPE. Our method produces better 3D geometry in all datasets.}
\label{tab:compare_baselines_human_geometry}
\makebox[\textwidth]{%
\begin{tabular}{l c c c c | c c c c | c c c c}
 \hline
& \multicolumn{4}{c}{Sizer Dataset~\cite{tiwari2020sizer}} & \multicolumn{4}{c}{IIIT Dataset~\cite{jinka2023iiit}} & \multicolumn{4}{c}{CAPE Dataset~\cite{jinka2023iiit}}\\
 \hline
Method & {$\text{CD}_\text{(cm)}$ $\downarrow$} & {$\text{P2S}_\text{(cm)}$ $\downarrow$} & {F-score $\uparrow$}   & $\text{NC} \uparrow$ & {$\text{CD}_\text{(cm)}$ $\downarrow$} & {$\text{P2S}_\text{(cm)}$ $\downarrow$} &  {F-score $\uparrow$}   & $\text{NC} \uparrow$ & {$\text{CD}_\text{(cm)}$ $\downarrow$} & {$\text{P2S}_\text{(cm)}$ $\downarrow$} & {F-score $\uparrow$}   & $\text{NC} \uparrow$\\
\hline
SMPL~\cite{loper2015smpl} &  $3.94$   & $4.02$ & $ 0.237$ & $0.743 $ & $4.67$ & $4.33$  &$0.204$  & $0.728$ & $5.04$  &$ 4.91$ & $0.213$ & $ 0.743$   \\
 \hline
PiFU~\cite{saito2019pifu}   &  $2.35$   & $2.31$ & $ 0.410$ & $0.782 $ & $2.70$ & $2.64$  &$0.337$  & $0.764$ & $3.40$  &$ 3.27$ & $0.314$ & $ 0.791$   \\
FoF~\cite{feng2022fof}     &  $5.37$   & $5.26$ & $0.204$ & $0.676$ & $5.34$ & $5.29$  & $0.188$  & $0.691$ & $5.65$  &$5.52$ & $0.146$ & $0.689$   \\
ICON~\cite{xiu2023icon}    &  $3.01$   & $3.20$ & $ 0.285$ & $0.771$ & $4.55 $ & $4.53$  &$0.202$  & $0.716$ & $4.28$  &$4.28$ & $0.238$ & $0.762$   \\
ECON~\cite{xiu2023econ}    &  $2.83$   & $3.04$ & $0.329$ & $0.781$ & $3.86$ & $3.84$  &$0.253$  & $0.744$ & $3.96$  &$4.14$ & $0.286$ & $0.775$   \\
SiTH~\cite{ho2023sith}   &  $3.38$   & $3.45$ & $0.285 $ & $0.753$ & $4.90$ & $4.83$  &$0.208$  & $0.716$ & $3.76$  &$ 3.95$ & $0.279$ & $0.785$    \\
SiFU~\cite{zhang2023sifu} &  $2.69$   & $2.81$ & $0.324$ & $0.778$ & $4.25$ & $4.18$  &$0.216$  & $0.725$ & $3.73$  &$3.71$ & $0.270$  & $0.779$   \\
$\text{LGM}_\text{ft}$~\cite{Tang2024LGM}  &  $2.80$   & $3.27$ & $0.306 $ & $0.556$ & $3.76$ & $4.31$  &$0.245$  & $0.567$ & $3.96$  &$4.23$ & $0.258$ & $0.557$    \\
\hline
 $\textit{Gen3D}_\text{avatar}$   &  $\textbf{1.06 }$   & $\textbf{ 1.05}$ & $\textbf{0.627}$ & $\textbf{0.794}$ & $\textbf{1.44}$ & $\textbf{1.39}$  &$\textbf{0.531}$  & $ \textbf{0.781}$ & $\textbf{1.89}$  &$\textbf{1.84}$ &$\textbf{0.491}$ &$\textbf{0.801}$ \\
\hline
\end{tabular}
}
\end{table*}

\begin{table*}[t!]
\caption{\textbf{Appearance evaluation for clothed avatars reconstruction} on Sizer, IIIT, and CAPE dataset. Our method produces overall better appearance.}
\label{tab:compare_baselines_human_appearance}
  \centering
\begin{tabular}{l c c c c | c c c c | c c c c}
 \hline
& \multicolumn{4}{c}{Sizer Dataset~\cite{tiwari2020sizer}} & \multicolumn{4}{c}{IIIT Dataset~\cite{jinka2023iiit}} & \multicolumn{4}{c}{CAPE Dataset~\cite{jinka2023iiit}}\\
 \hline
Method & {PSNR $\uparrow$}  & {SSIM $\uparrow$} & { LPIPS $\downarrow$} &{FID $\downarrow$} & {PSNR $\uparrow$}  & {SSIM $\uparrow$} & { LPIPS $\downarrow$} &{FID $\downarrow$} & {PSNR $\uparrow$}  & {SSIM $\uparrow$} & { LPIPS $\downarrow$} &{FID $\downarrow$}\\
 \hline
PiFU~\cite{saito2019pifu}   &  $19.22$   & $0.913$ & $ 0.068$ & $33.50$ & $22.40$ & $\textbf{0.905}$  &$0.083$  & $22.41$ & $22.03$  &$ 0.910$ & $0.082$ & $ 38.79 $   \\
SiTH~\cite{ho2023sith}   &  $18.90$   & $0.912$ & $0.063 $ & $21.87$ & $19.53$ & $0.901$  &$0.078$  & $ 19.90$ & $22.20$  &$ 0.908$ & $0.082$ & $28.46$    \\
SiFU~\cite{zhang2023sifu} &  $18.01$   & $0.899$ & $0.072$   &$36.64$ & $\textbf{22.65}$ & $0.899$ & $0.087$ & $46.76$ & $\textbf{22.23}$  &$0.906$ & $0.085$  & $43.63$   \\
$\text{LGM}_\text{ft}$~\cite{Tang2024LGM}  &  $20.57$   & $0.902$ & $0.077 $ & $16.75$ & $20.65$ & $0.879$  &$0.100$  & $15.54$ & $20.46$  &$0.898$ & $0.089$ & $20.33$    \\
\hline
 $\textit{Gen3D}_\text{avatar}$    &  $\textbf{21.28}$   & $\textbf{0.928}$ & $\textbf{0.047}$ & $\textbf{10.01}$ & $22.13$ & $\textbf{0.905}$  &$\textbf{0.066}$  & $ \textbf{9.69}$ & $21.46$  &$\textbf{0.912}$ &$\textbf{0.064}$ &$\textbf{16.40}$ \\
\hline
\end{tabular}
\end{table*}

\begin{figure*}[ht]
    \centering
    \begin{minipage}{0.4\textwidth}
        \centering
        \captionof{table}{\textbf{Runtime performance comparison}. Our method is faster than template-based human reconstruction methods.}
        \label{tab:inference_efficiency_avatar}
        \footnotesize
        \begin{tabular}{ l c c }
        \hline
         &   Time (s) & VRAM (GB)  \\     
         \hline
        ICON~\cite{xiu2023icon} &  $60.5$ & $6.3$   \\
        ECON~\cite{xiu2023econ} &  $45.3$ & $5.9$ \\
        SiFU~\cite{zhang2023sifu} & $48.9$ & $12.0$ \\
        SiTH~\cite{ho2023sith} &  $106.2$ & $22.0 $ \\
        \hline
        $\textit{Gen3D}_\text{avatar}$ & $22.6$ & $11.7$  \\
         \hline
        \end{tabular}
    \end{minipage}%
    \hfill
    \begin{minipage}{0.6\textwidth}
        \centering
        \includegraphics[width=0.7\textwidth]{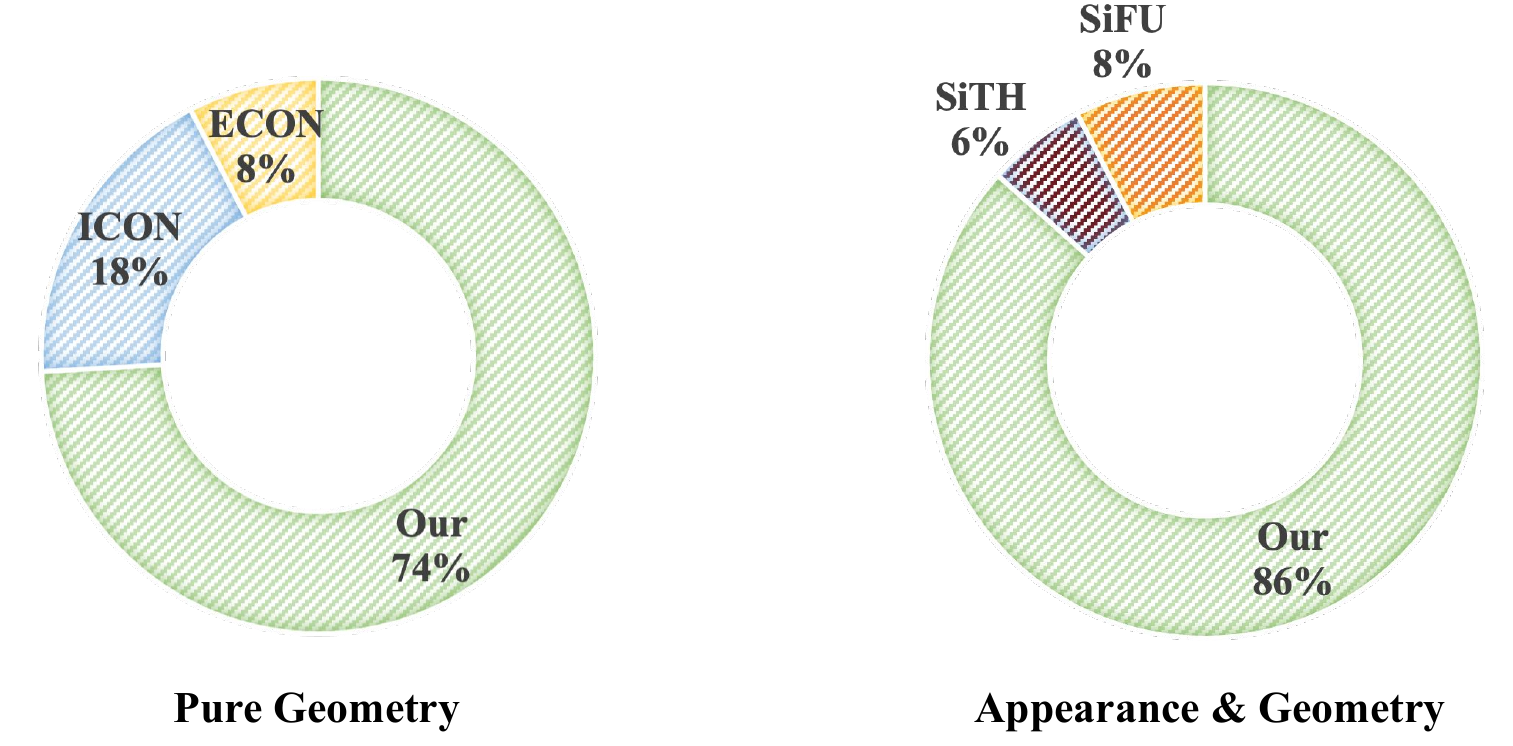}  
    \end{minipage}%
    \caption{\textbf{Runtime performance \& user preference comparison for avatars reconstruction.} Left: Inference time and GPU consumption of SoTA avatar reconstruction approaches. Right: User study statistics of avatar reconstruction on pure geometry or appearance \& geometry comparison. Our method is preferred by most people in both geometry and appearance.}
    \label{fig:user-study}
\end{figure*}

\begin{figure*}[t!]
  \centering
  \includegraphics[width=\textwidth]{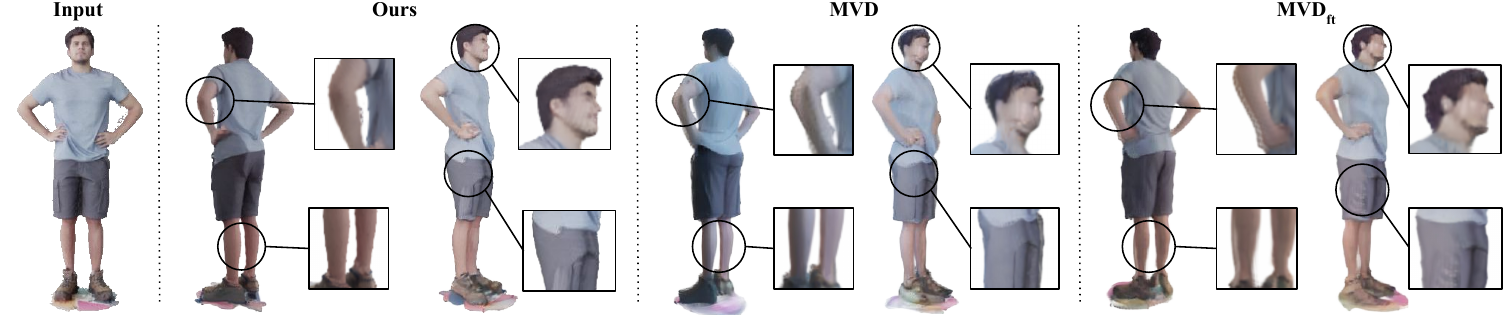}
  \vspace{-0.5cm}
  \caption{\textbf{3D reconstruction conditioned on different multi-view priors.} Without our 3D-consistent sampling, the 2D diffusion model cannot generate 3D consistent multi-views (MVD, $\text{MVD}_\text{ft}$), leading to artifacts like floating 3D Gaussian splats. Our method obtains more consistent multi-views hence better 3D-GS and rendering. }
  \label{fig:ablate_consistent_diff}
\end{figure*}

We evaluate $\textit{Gen3D}_\text{avatar}$ in novel view synthesis and the geometry on the Sizer dataset~\cite{tiwari2020sizer}, IIIT\cite{jinka2023iiit}, and CAPE dataset~\cite{zhang2017buff, ponsmoll2017clothcap, ma2019cape}.
We compare our approach against prior methods for image-to-avatar reconstruction, including pure geometry-based~\cite{feng2022fof, xiu2023icon, xiu2023econ} and textured geometry-based~\cite{saito2019pifu, zhang2023sifu, ho2023sith} human reconstruction methods. To further assess performance, we also fine-tuned the state-of-the-art object reconstruction method LGM~\cite{Tang2024LGM} and its underlying multi-view diffusion model~\cite{Wang2023ImageDream} on our training data, denoted as $\text{LGM}_{\text{ft}}$. 

We report quantitative evaluation in~\cref{tab:compare_baselines_human_geometry} and~\cref{tab:compare_baselines_human_appearance} and show qualitative comparison for both appearance and geometry via extracted mesh in~\cref{fig:human_comparison}. It can be seen that template-based approaches heavily rely on accurate SMPL estimations hence they easily fail when the estimations are off. This is common when the person is wearing large dress, has a different shape as the adult SMPL body shape or is interacting with object/accessories. In contrast, our method is template-free hence can flexibly represent all possible body and clothing shapes, leading to more coherent appearance and geometry reconstruction. 


To further evaluate the reconstruction quality, we conduct a user study to compare the reconstruction of different methods. We render 20 subjects with texture to compare ours against SiTH and SIFU and another 20 subjects with only geometry to compare ours against ICON and ECON. The subjects are randomly sampled from evaluation dataset of Sizer~\cite{tiwari2020sizer}, IIIT~\cite{jinka2023iiit}, CAPE~\cite{ma2019cape}. We release the user study to 70 people from different technical backgrounds. Overall, our results are preferred by $80.3\%$ of users. It clearly shows that our $\textit{Gen3D}_\text{avatar}$ significantly outperforms baselines in both geometry and appearance. Please see \cref{fig:user-study} for visualization of the user study results. 

Our method is a diffusion-based feed-forward approach without any SMPL estimation or test-time optimization process as is typical in template-based methods~\cite{xiu2023icon, xiu2023econ, ho2023sith, zhang2023sifu}. This allows us to obtain 3D reconstruction at higher inference speed. We report the runtime comparison (on Nvidia A100) in~\cref{tab:inference_efficiency_avatar}. Our method is much faster than baseline human reconstruction methods.

\subsection{Ablation Studies}
\label{subsec:ablation}
In this section, we elaborate our ablation studies which validate our design choices. For \textit{3D Diffusion helps 2D Diffusion} and \textit{2D Diffusion helps 3D Diffusion}, we focus on our human model $\textit{Gen3D}_{\text{avatar}}$ and report results on the human datasets as well as generalization to o.o.d unseen objects. We also evaluate the robustness of our $\textit{Gen3D}_{\text{obj}}$ model to diverse input viewpoints, especially non-zero elevation angles that frequently occur in real-world object captures.

\vspace{0.2cm}
\noindent\textbf{3D Diffusion helps 2D Diffusion.} 
One of our key ideas is leveraging our explicit 3D model to refine the 2D multi-view reverse sampling trajectory, ensuring 3D consistency in 2D Multi-View Diffusion (MVD) generation (see~\cref{sec:3dhelps2d} and ~\cref{eq:xt_minus_one_from_xt_x0_ours}).
To evaluate this, we compare the multi-view images (4 orthogonal views) generated by pretrained MVD~\cite{Wang2023ImageDream}, fine-tuned MVD on our human data ($\text{MVD}_\text{ft}$) and MVD with our 3D consistent sampling (ours), as shown in \cref{tab:ablate_traj_refine_for_2d}. The results demonstrate that our proposed method effectively enhances the quality of generated multi-view images by leveraging the explicit 3D model to refine sampling trajectory.

Additionally. we analyze the 3D reconstruction results with the multi-view images generated by these models in \cref{fig:ablate_consistent_diff}. MVD and $\text{MVD}_{\text{ft}}$ produce inconsistent multi-view images, which typically lead to floating Gaussian and hence blurry boundaries. In contrast, our method can generate more consistent multi-views, result in better 3D Gaussian Splats and sharper renderings. 
We further quantitatively evaluate the impact of our proposed sampling trajectory refinement on final 3D reconstruction in~\cref{tab:ablate_traj_refine_for_3d}. We compare the reconstruction results of methods with and without our trajectory refinement while using the same 2D MVD and 3D reconstruction models with same setting in~\cref{tab:compare_baselines_human_appearance} and~\cref{tab:compare_baselines_human_geometry}. It can be clearly seen that our trajectory refinement improves the quality of 3D reconstruction.

\vspace{0.2cm}
\noindent\textbf{2D Diffusion helps 3D Diffusion.} Another key idea of our work is the use of multi-view priors $\tilde{\mathbf{x}}_0^\text{tgt}$ from 2D diffusion model pretrained on massive data~\cite{Deitke2023Objaverse, Rombach2022StableDiffusion, Schumann2022Laion5B} to enhance our 3D generative model. This additional prior information is pivotal for ensuring accurate reconstruction of both in-distribution human dataset and generalizing to out-of-distribution objects. 

We evaluate the performance of our 3D model $g_{\phi}$ by comparing generation results with and without the 2D diffusion prior $\tilde{\mathbf{x}}_{0}^{\text{tgt}}$ (refer to \cref{eq:diffusion_with_x0_clean} and \cref{eq:diffusion_wo_x0}).
For avatars reconstruction, our powerful 3D reconstruction model can already achieve state-of-the-art performance. Moreover, our $\textit{Gen3D}_{\text{avatar}}$ full model with multi-view prior $\tilde{\mathbf{x}}_0^\text{tgt}$ generates avatars with higher quality as demonstrated in \cref{tab:ablate_multiview_cond_human}. We further evaluate it on the GSO~\cite{downs2022gso} dataset which consists of unseen general objects to our $\textit{Gen3D}_{\text{avatar}}$ model. 
The improvements are even more pronounced in this setting, highlighting the challenges of generating coherent 3D structures from a single 2D image, particularly with unseen objects. These ablation studies effectively proves that the 2D multi-view diffusion prior enhances generalization capability.

\textbf{Robustness to input pose variations.} To systematically measure robustness, we render synthetic input images by varying the input elevation angles from 0° (frontal) to 30° (overhead) while keeping azimuth angle fixed.  We report the standard 2D metrics (PSNR, SSIM, LPIPS, FID) and 3D scores (Chamfer, P2S, NC, F-score)in \Cref{tab:compare_elevation_object}. Over this 0°–30° range, PSNR varies by only ±0.40 dB and SSIM by ±0.006 (mean ± std), confirming that Gen-3Diffusion generalizes robustly to the camera elevation angles encountered in everyday object-centric photography.
\section{Overview}
\label{sec:conclusion}

\begin{table*}[h!]
\caption{\textbf{Ablating the influence of 3D consistent guided sampling} for 3D-GS generation. Our proposed sampling strategy improves the 3D reconstruction quality by enhancing multi-view consistency of 2D diffusion models.}
\label{tab:ablate_traj_refine_for_3d}
\centering
\footnotesize
\begin{tabular}{ l c c c c c c }
\hline
{Method} & {$\text{CD}_\text{(cm)}$$\downarrow$} & {F-score$\uparrow$}   & $\text{NC}\uparrow$ & { LPIPS$\downarrow$} & {SSIM$\uparrow$}  &{PSNR$\uparrow$}  \\     
\hline
Our w/o Traj. Ref. & $1.57$ & $0.498$ & $0.794$ & $0.064$ & $0.908 $ &  $21.09$\\
\hline
 \emph{Ours} &  $\textbf{1.35}$ &  $\textbf{0.550}$ &  $\textbf{0.798}$ & $\textbf{0.060}$ & $\textbf{0.918}$ & $\textbf{21.49}$ \\
 \hline
\end{tabular}
\end{table*}

\begin{table*}[h!]
\caption{\textbf{Ablating the influence of 2D multi-view priors $ \tilde{\mathbf{x}}^{\text{tgt}}_{0}$}. The strong prior from 2D diffusion models enhance the 3D reconstruction quality.}
\label{tab:ablate_multiview_cond_human}
\centering
\footnotesize
\begin{tabular}{ l c c c c c c }
\hline
{Method} & {$\text{CD}_\text{(cm)}$$\downarrow$} & {F-score$\uparrow$}   & $\text{NC}\uparrow$ & { LPIPS$\downarrow$} & {SSIM$\uparrow$}  &{PSNR$\uparrow$}  \\     
\hline
Ours w/o $ \tilde{\mathbf{x}}^{\text{tgt}}_{0}$ & $1.75$ & $0.498$ & $0.795$ & $0.068$ & $0.912 $ &  $20.98$
 \\
\hline
 \emph{Ours}  &  $\textbf{1.35}$ &  $\textbf{0.550}$ &  $\textbf{0.798}$ & $\textbf{0.060}$ & $\textbf{0.918}$ & $\textbf{21.49}$ 
 \\
 \hline
\end{tabular}
\end{table*}

 \begin{table*}[h!]
\centering
   \caption{\textbf{Performance of Gen-3Diffusion under different input elevations.} Our performance only varies slightly (PSNR ±0.4 dB and SSIM ±0.006) when input elevation changes from 0 to 30 degrees, showing that our Gen-3Diffusion is robust to different daily camera elevation settings.
   }
\label{tab:compare_elevation_object}
\begin{tabular}{ c c c c c c c c c}
 \hline
Elevation  &{PSNR $\uparrow$}  & {SSIM $\uparrow$} & { LPIPS $\downarrow$} &{FID $\downarrow$} & {$\text{CD}_\text{(cm)}$ $\downarrow$}& {$\text{P2S}_\text{(cm)}$ $\downarrow$} &  $\text{NC} \uparrow$ & {F-score $\uparrow$}    \\
 \hline
 0 & $22.30$ & $0.911$ & $0.083$  &$ \textbf{59.08} $ &  $4.28$  &  $ 3.98 $ & $\textbf{0.723}$  &  $0.231$\\
 5 & $ 22.57 $ & $ \textbf{0.917} $ & $ 0.082 $  &$ 62.06 $  &  $4.62$    &  $4.18 $   & $ 0.729 $ & $ 0.243 $  \\
 10 & $22.54$ & $0.911$ & $0.081$  &$ 60.67 $  & $ 3.71 $  &  $3.16$    & $ 0.736 $  &$ 0.240 $ \\
 15 & $\textbf{22.61}$ & $0.912$ & $\textbf{0.078}$  &$ 59.56 $  & $ \textbf{3.22} $   & $ \textbf{2.96} $ &  $0.746$    &$ \textbf{0.260} $ \\
 20 & $ 22.31 $ & $0.910$ & $0.081$  &$ 61.32 $   &  $3.53$  &  $ 3.27 $    & $0.741$  & $0.254$ \\
25  & $ 22.11 $ & $ 0.908 $ & $ 0.084 $  &$ 61.89 $  &  $ 3.75 $  &  $ 3.41 $    & $ 0.733 $ &  $0.247$  \\
30 & $ 21.46 $ & $ 0.901 $ & $0.087$   &$ 65.46 $   &  $ 3.76 $   &  $ 3.78$   & $ 0.731 $  &   $0.242$ \\
 \hline
Mean $\pm$ Std & $\textbf{22.27} \pm 0.40 $ & $ \textbf{0.910} \pm 0.005 $ & $\textbf{0.082} \pm 0.003$   &$ \textbf{61.43} \pm 2.10 $   &  $ \textbf{3.84} \pm 0.47 $   &  $ \textbf{3.53}\pm 0.45$   & $ \textbf{0.734} \pm 0.008 $  &   $\textbf{0.245} \pm 0.010$ \\
 \hline
\end{tabular}
\end{table*}

\begin{table}[t!]
\caption{\textbf{Ablating the influence of 3D consistent guided sampling} for 2D multi-view images generation. Our proposed sampling strategy improves the multi-view image quality from 2D diffusion models. 
}
\label{tab:ablate_traj_refine_for_2d}
\centering
\footnotesize
\begin{tabular}{ l c c c }
\hline
{Method} &{PSNR $\uparrow$}  & { LPIPS $\downarrow$} & { SSIM $\uparrow$}  \\     
 \hline
MVD & $22.32$  & $0.078$ & $0.911$   \\
$\text{MVD}_\text{ft}$ &  $24.14$ & $0.061$ & $0.926 $ \\
\hline
 \emph{Ours} & $\textbf{24.69}$ &  $\textbf{0.048}$ & $\textbf{0.934}$  \\
 \hline
\end{tabular}
\end{table}

\subsection{Limitations}
Limited by low resolution ($256\times256$) of our adopted pretrained 2D diffusion model~\cite{Wang2023ImageDream}, our model cannot recover fine details such as detailed human face. We noticed that the small VAE space ($32\times32$) could be the bottleneck for representing the face of human and detailed textures of objects.  Extending the current framework to higher resolution could be the key for enhancing reconstruction fidelity.

\subsection{Future Works}
Our \textbf{Gen-3Diffusion} is a general framework for image-to-3D reconstruction, which shows the superior reconstruction ability on isolated objects and human subjects. Extending the current framework to scene-level reconstruction yields more difficulties such as  different camera poses and z-buffering. We leave this for future works.

\subsection{Conclusion}
In this paper, we introduce \textbf{Gen-3Diffusion}, a 3D consistent diffusion model for creating realistic 3D objects or clothed avatars from single RGB images. 
Our key ideas are two folds: 1) Leveraging strong multi-view priors from pretrained 2D diffusion models to generate 3D Gaussian Splats, and 2) Using the reconstructed explicit 3D Gaussian Splats to refine the sampling trajectory of the 2D diffusion model which enhances 3D consistency. 
We carefully designed a diffusion process that synergistically combines the strengths of both 2D and 3D models.
We compare our image-to-3D model $\textit{Gen3D}_\text{object}$ with 9 state-of-the-art methods and our image-to-avatar model $\textit{Gen3D}_\text{avatar}$ with 6 popular works, show that our approach outperforms them in both appearance and geometry. 
We also extensively ablate our method which proves the effectiveness of our proposed ideas. 
Our code and pretrained models will be publicly available on our \href{https://yuxuan-xue.com/gen-3diffusion}{project page}.

\begin{figure}[t!]
  \centering
  \includegraphics[width=0.45\textwidth]{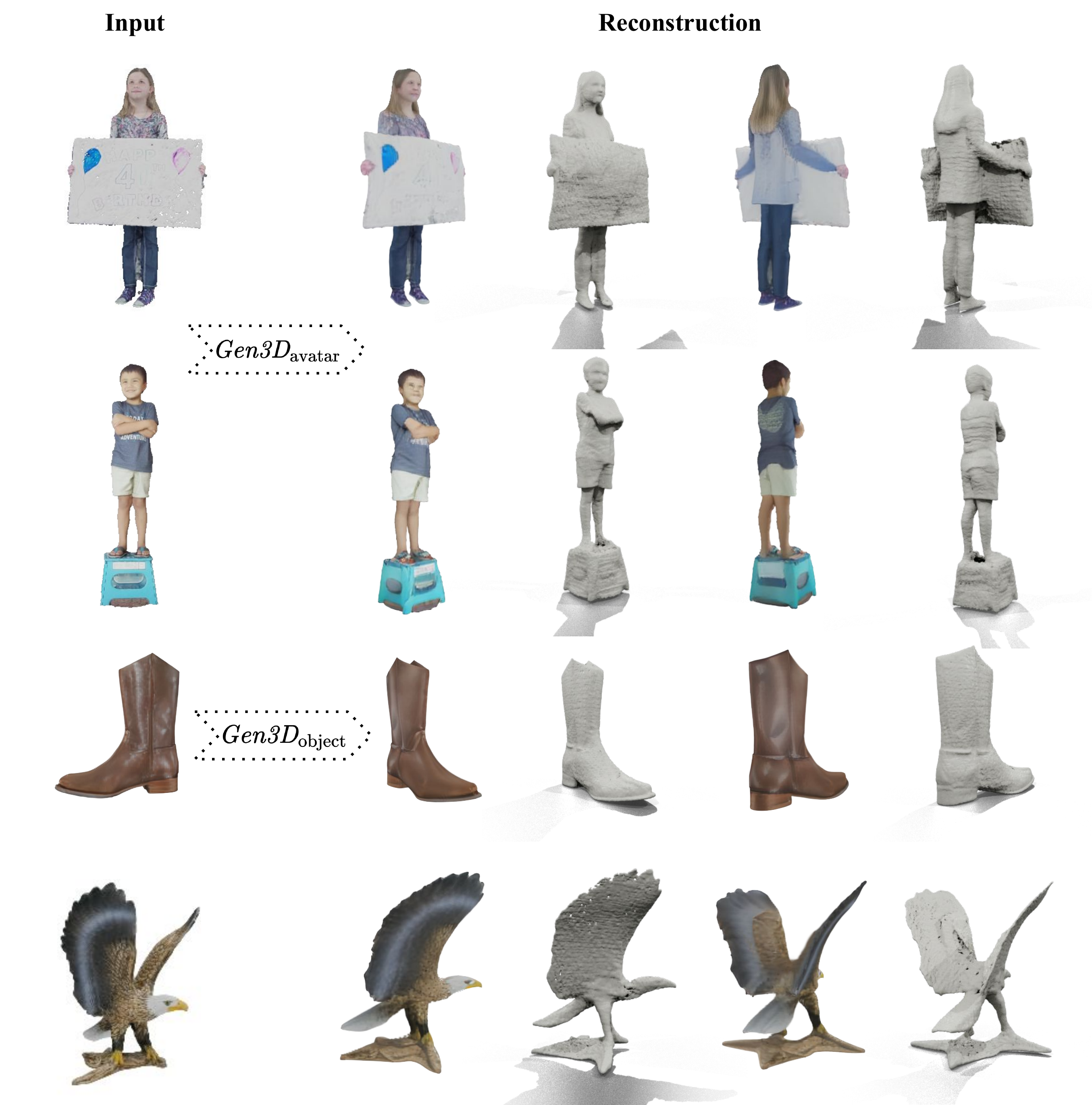}
  \vspace{0cm}
  \caption{Gallery of 3D reconstruction from single RGB images using our $\textit{Gen3D}_\text{avatar}$ and $\textit{Gen3D}_\text{object}$.}
  \label{fig:gallery}
\end{figure}


%

\ifCLASSOPTIONcompsoc
  \section*{Acknowledgments}
\else
  \section*{Acknowledgment}
\fi

This work is made possible by funding from the Carl Zeiss Foundation. 
This work is also funded by the Deutsche Forschungsgemeinschaft (DFG, German Research Foundation) - 409792180 (EmmyNoether Programme, project: Real Virtual Humans) and the German Federal Ministry of Education and Research (BMBF): Tübingen AI Center, FKZ: 01IS18039A. 
The authors thank the International Max Planck Research School for Intelligent Systems (IMPRS-IS) for supporting Y.Xue.
R. Marin has been supported by the innovation program under Marie Skłodowska-Curie grant agreement No 101109330.
G. Pons-Moll is a member of the Machine Learning Cluster of Excellence, EXC number 2064/1 – Project number 390727645.

\ifCLASSOPTIONcaptionsoff
  \newpage
\fi

{\small
    \bibliographystyle{IEEEtran}
    \bibliography{literatures}
}
 


%

\begin{IEEEbiography}[{\includegraphics[width=1in,height=1.25in,clip,keepaspectratio]{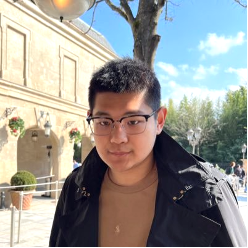}}]{Yuxuan Xue}
is currently pursuing Ph.D. degree in the Real Virtual Human (RVH) group at the Univerisity T\"ubingen, supervised by Prof. Dr. Gerard Pons-Moll. Before that, he received B.Sc. (2016) and M.Sc. (2022) from Technical University of Munich (TUM). His research interests lie on perceiving human from real world and modelling into metaverse. He has published at top conferences and journal in machine learning and vision (ICCV, NeurIPS, ICLR, IJCV). His work got Best Student Paper Award at BMVC 2022.
\end{IEEEbiography}

\begin{IEEEbiography}[{\includegraphics[width=1in,height=1.25in,clip,keepaspectratio]{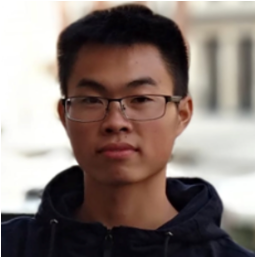}}]{Xianghui Xie}
is currently pursuing Ph.D. degree in the Real Virtual Human (RVH) group at the Univerisity T\"ubingen and Max Planck Institute for Informatics, supervised by Prof. Dr. Gerard Pons-Moll. Before that, he obtained B.Sc. from KU Leuven and M.Sc from Saarland university . His research interests lie on modelling interaction between human and objects. He has published at top machine learning and vision conferences (ECCV, CVPR, NeurIPS). 
\end{IEEEbiography}


\begin{IEEEbiography}
[{\includegraphics[width=1in,height=1.25in,clip,keepaspectratio]{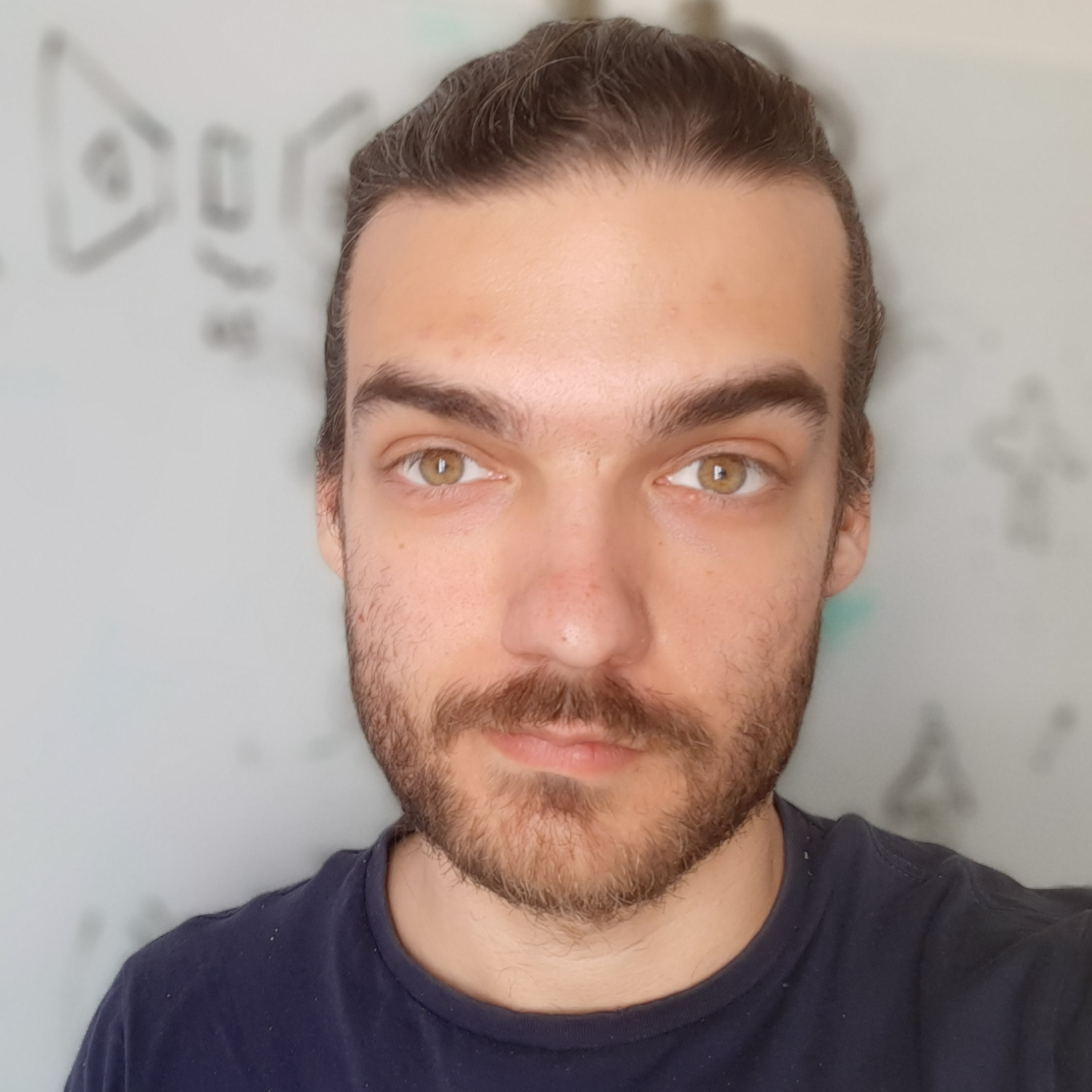}}]{Riccardo Marin} received the PhD degree in Computer Science from the University of Verona, Italy, in 2021. After that, he was a post-doc researcher and Adjunct Professor at the Sapienza University of Rome as part of the GLADIA lab.
In 2022, he joined the University of Tuebingen as a post-doc researcher in the Real Virtual Humans group, funded by a Humboldt Foundation Research Fellowship and a Marie Skłodowska-Curie Post-Doctoral Fellowship.
He is currently a post-doc researcher at the Technical University of Munich (TUM) as part of the Computer Vision Group.
His research interests include 3D Shape Analysis, Matching and Registration, Geometric Deep Learning, and Virtual Humans.
\end{IEEEbiography}

\begin{IEEEbiography}[{\includegraphics[width=1in,height=1.25in,clip,keepaspectratio]{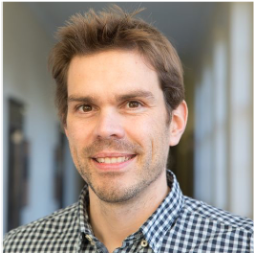}}]{Gerard Pons-Moll}
is Professor at the University
of Tuebingen, head of the Emmy Noether inde-
pendent research group ”Real Virtual Humans”,
senior researcher at the Max Planck for Infor-
matics (MPII) in Saarbrucken, Germany. His re-
search lies at the intersection of computer vision,
computer graphics and machine learning – with
special focus on analyzing people in videos, and
creating virtual human models by ”looking” at
real ones. His work has received several awards
including the prestigious Emmy Noether Grant
(2018), a Google Faculty Research Award (2019), a Facebook Reality
Labs Faculty Award (2018), and the German Pattern Recognition Award
(2019). His work got Best Papers Awards and nomminations at CVPR’20,
CVPR’21, ECCV’22. He serves regularly as area chair for the top
conferences in vision and graphics (CVPR, ICCV, ECCV, Siggraph).
\end{IEEEbiography}



\end{document}